\documentclass{article}

\usepackage[preprint,nonatbib]{neurips_2024}

\usepackage[utf8]{inputenc} % allow utf-8 input
\usepackage[T1]{fontenc}    % use 8-bit T1 fonts
\usepackage{hyperref}       % hyperlinks
\usepackage{url}            % simple URL typesetting
\usepackage{booktabs}       % professional-quality tables
\usepackage{amsfonts}       % blackboard math symbols
\usepackage{nicefrac}       % compact symbols for 1/2, etc.
\usepackage{microtype}      % microtypography
\usepackage{xcolor}         % colors

\usepackage{amsmath,amssymb,graphicx}
\usepackage{lipsum} % Only used to generate random text.
\usepackage{amsmath}
\usepackage{xcolor}

\usepackage{booktabs}

\title{HDNet: Physics-Inspired Neural Network for Flow Estimation
  based on Helmholtz Decomposition}

\author{%
  Miao Qi \\
  Visual Computing Center\\
  King Abdullah University of Science and Technology\\
  Thuwal, Saudi Arabia\\
  \texttt{miao.qi@kaust.edu.sa} \\
  \and
  Ramzi Idoughi \\
  Visual Computing Center\\
  King Abdullah University of Science and Technology\\
  Thuwal, Saudi Arabia\\
  \texttt{ramzi.idoughi@kaust.edu.sa} \\
  \and
  Wolfgang Heidrich \\
  Visual Computing Center\\
  King Abdullah University of Science and Technology\\
  Thuwal, Saudi Arabia\\
  \texttt{wolfgang.heidrich@kaust.edu.sa} \\
  \and
}

\newcommand{\etal}{\textit{et al.}}

\newcommand{\rVector}{\mathbf{r}}

\newcommand{\ReflectiveIndex}{p}

\newcommand{\VelocityField}{v}

\newcommand{\nnweight}{\mathbf{\xi}}
\DeclareMathOperator*{\argmin}{arg\,min}

\newcommand{\frepara}{\omega_0}

\newcommand{\vv}{\mathbf{v}}
\newcommand{\vgen}{\mathbf{v}^*}
\newcommand{\vsol}{\mathbf{v_{sol}}}
\newcommand{\virr}{\mathbf{v_{irr}}}
\newcommand{\divv}{\nabla\cdot}
\newcommand{\curl}{\nabla\times}
\newcommand{\laplac}{\nabla^2}
\newcommand{\potential}{\phi}

\newcommand{\Velocityfirstcomponent}{u}
\newcommand{\Velocitysecondcomponent}{w}

\newcommand{\ScalarPotential}{\phi}
\newcommand{\VectorPotential}{\Phi}

\newcommand{\lossweight}{\lambda_1}
\newcommand{\lossgradweight}{\lambda_2}
\newcommand{\ProjHDNet}{\mathcal{D}}

\newcommand{\MLP}{f}
\newcommand{\xaxis}{\ensuremath{x}}  % x-axis
\newcommand{\yaxis}{\ensuremath{y}}  % y-axis
\newcommand{\taxis}{\ensuremath{t}}  % t-axis

\begin{document}

\maketitle

\begin{abstract}
Flow estimation problems are ubiquitous in scientific imaging. Often, the underlying flows are subject to physical constraints that can be exploited in the flow estimation; for example, incompressible (divergence-free) flows are expected for many fluid experiments, while irrotational (curl-free) flows arise in the analysis of optical distortions and wavefront sensing. In this work, we propose a Physics-Inspired Neural Network (PINN) named HDNet, which performs a Helmholtz decomposition of an arbitrary flow field, i.e., it decomposes the input flow into a divergence-only and a curl-only component. HDNet can be trained exclusively on synthetic data generated by reverse Helmholtz decomposition, which we call Helmholtz synthesis. As a PINN, HDNet is fully differentiable and can easily be integrated into arbitrary flow estimation problems.
\end{abstract}

\section{Introduction}

In many flow estimation problems, the reconstructed flows are governed by physical properties. For example, incompressibility (divergence-free) holds for many flows in fluid simulations and experiments, while irrotationality (curl-free) of optical flows is expected in some applications of optical distortion analysis. Incorporating these physical constraints into the reconstruction framework can significantly improve the accuracy of the results in such inverse problems, as proven by several existing works~\cite{gregson2014capture,xiong2017rainbow,xiong2018reconfigurable,xiong2020rainbowpiv,zang2020tomofluid,chen2021snapshot,du2021study}. However, enforcing these constraints in a differentiable manner, compatible with popular deep learning reconstruction methods, remains a challenging problem. 
A straightforward approach involves incorporating physical laws as loss terms of the neural network pipeline~\cite{caiphysics,raissi2020hidden,molnar2023estimating}. This approach is better known as Physics-informed Neural Networks (PINNs). For instance, in the case of incompressibility and irrotationality, an $ \ell_2$ norm of the divergence or the curl would be added to the total loss. These soft constraints are easy to implement but do not guarantee that the reconstruction aligns perfectly with the physical constraints, as illustrated in Fig.~\ref{PIV2Fig:RealPIV}. Enforcing physical properties as a hard constraint has already been proposed for the incompressibility of the flow through the use of the pressure projection~\cite{gregson2014capture,xiong2017rainbow,xiong2020rainbowpiv,qi2023scattering}. However, this technique suffers from its lack of differentiability and its incompatibility with popular deep learning-based reconstruction methods.

To address this challenge, we propose a novel approach based on the
\emph{fundamental theorem of vector calculus} also known as
\emph{Helmholtz decomposition}. Specifically, we introduce the
Helmholtz Decomposition Network (HDNet), which is a flexible and
differentiable network that enforces physical constraints during flow
reconstructions. HDNet decomposes any flow field into two components:
a solenoidal (incompressible) flow field and an irrotational flow
field. In addition to these two vector fields, HDNet also outputs the
scalar potential of the irrotational field component, which has
different physical interpretations according to the application
context. For example, this scalar potential corresponds to the
normalized pressure in fluid dynamics applications, while it
represents the phase profile in distortion problems like
Background-Oriented Schlieren (BOS) imaging and wavefront sensing (see
Section~\ref{PIV2Sec:HDMath}).

To enable a supervised training of HDNet, it is necessary to generate a large-scale dataset of paired input-output data. Conventional fluid simulation software, however, is computationally expensive and time-consuming, hindering the generation of such datasets. Therefore, we propose the Helmholtz synthesis module to generate efficiently a large-scale fluid training dataset. This module, based on the reverse process of Helmholtz decomposition, enables the creation of large-scale, highly variable fluid data pairs in a short timeframe, which makes supervised learning for HDNet possible. 

Furthermore, we propose a PINN-based flow reconstruction pipeline (Fig.~\ref{PIV2Fig:teaser} (b)), that leverages HDNet to enforce physical constraints. This pipeline is demonstrated in the context of incompressible flow in fluid dynamics applications. We also illustrate two examples of irrotational flow in the context of optical flow estimation in phase distortion problems. With an extensive comparative study, we show that our method outperforms the conventional Horn-Schunk optical flow and soft constraint methods in terms of reconstruction accuracy and preservation of physical properties.

Our HDNet exhibits significant flexibility beyond its application in our PINN-based flow reconstruction pipeline. Its capabilities extend to a wide range of deep learning applications, including inverse imaging pipelines, differential reconstruction frameworks, and even forward simulation problems.

In summary, our contributions are:

\begin{itemize}

\item We propose HDNet, a differentiable network to enforce the physical constraints for flow reconstruction problems. HDNet is highly versatile and capable of simultaneously obtaining solenoidal, irrotational fields, and scalar potential fields at the same time.

\item We propose Helmholtz synthesis, an efficient data generation method capable of creating large-scale paired datasets for the training of HDNet.

\item As an example application, we demonstrate how to integrate HDNet into a fully differentiable PINN-based flow reconstruction pipeline, resulting in exceptional reconstruction performance while rigorously preserving physical constraints.

\item We evaluate our approach on different real applications and show improvement in comparison to existing methods.
\end{itemize}

\section{Related Work}
\medskip\noindent
{\bfseries Physics-informed learning.} Physics-Informed Learning~\cite{karniadakis2021physics, kadambi2023incorporating} is a series of strategies that leverage physical laws and constraints to improve machine learning models' predictions. This approach has applications in several domains, including fluid dynamics, quantum mechanics, electromagnetics, and biology~\cite{caiphysics,zhang2018deep,bennini2022pinns,lagergren2020biologically}. One popular strategy involves the use of Physics-informed Neural Networks (PINNs). Firstly introduced by Raissi~\etal~\cite{raissi2019physics}, PINNs incorporate physical equations directly into the loss function, ensuring that the network's outputs align with the governing physical principles. PINNs have been extensively used in the field of fluid dynamics. For example, Cai~\etal~\cite{cai2021flow} employed a PINN to predict the pressure and the velocity field from the Tomographic background-oriented Schlieren temperature field for the flow over an espresso cup. Wang~\etal~\cite{wang2022dense} proposed a PINN to estimate the velocity field of fluids in microchannels. By their design, PINNs do not incorporate a physical forward model in their reconstruction process but rely only on integrating the physical equations in the form of soft constraints within the loss function. Another strategy consists of combining neural models with traditional physics-based simulations to incorporate physics through generated datasets~\cite{bear2021physion,ummenhofer2019lagrangian}. Additionally, some existing methods embedded physical priors or constraints into the network architecture. For instance, Cao~\etal~\cite{cao2022dynamic} propose a neural space-time model for representing dynamic samples captured using speckle structured illumination. Specifically, their approach utilizes two Multi-Layer Perceptrons (MLPs): one to model the motion field and another to represent a canonical configuration of the dynamic sample. The reconstructed sample at different times is obtained by warping the canonical scene with the estimated motion field. The reconstructed frames are then processed through a forward model simulating the imaging process, and the resulting outputs are compared to the captured images to compute the training loss.

\medskip\noindent
{\bfseries Enforcing incompressibility in fluid simulation and reconstruction.}
Fluid simulation and reconstruction tasks often require enforcing incompressibility as a hard constraint, which ensures the conservation of the fluid volume. This constraint is equivalent to having a divergence-free flow. In the literature, several conventional methods have been proposed to enforce this physical constraint. PCISPH~\cite{solenthaler2009predictive} corrects pressure terms in the Smoothed Particle Hydrodynamics (SPH) representation by assuming constant density. Vortex methods~\cite{park2005vortex,mimeau2021review,mimeau2021review} compute velocity from vortex strength using the Biot-Savart formula, facilitating incompressible fluid simulation. The pressure projection~\cite{gregson2014capture,qi2023scattering,xiong2017rainbow} approach decomposes an arbitrary field into solenoidal (divergence-free) and irrotational (curl-free) components. This involves solving the Poisson equation, often using iterative gradient-based methods like Preconditioned Conjugate Gradient (PCG). The solenoidal component will be selected to satisfy the incompressibility constraint. While these methods are effective, they share a common limitation: they are non-differentiable. Thus, they cannot be integrated into popular differentiable and deep learning pipelines. To address this challenge, recent research has explored differentiable physical constraint methods. A first approach involves applying a divergence or curl as a penalty term~\cite{caiphysics,raissi2020hidden,molnar2023estimating}, offering a simple and forward differentiable approach. However, the soft nature of these constraints may not always guarantee strict incompressibility. Moreover, some works~\cite{dong2019adaptive,tompson2017accelerating,ajuria2020towards} propose a Convolutional Neural Network (CNN) Poisson solver to replace the iterative solver in pressure projection. However, the lack of sufficient training datasets has led these approaches towards unsupervised learning, which may reduce their performance.

\medskip\noindent
{\bfseries Particle Image Velocimetry (PIV).}
PIV is a powerful imaging technique widely used for measuring fluid flow velocities in various fields~\cite{adrian2011particle,raffel2018particle}. The fundamental principle of PIV involves seeding the studied fluid or gas flow with particles. By illuminating the region of interest and recording the particles' advected motion, the fluid flows can be retrieved. In basic PIV techniques, the region of interest of the fluid is simply a plane illuminated with a laser light sheet and captured from a single camera, which leads to the reconstruction of an in-plane 2D velocity field. In this work, we demonstrate our approach using data captured with a basic 2D PIV approach.
% While the basic PIV technique is limited to retrieving in-plane 2D velocity fields, several approaches have been proposed to extend its performance. Stereo-PIV~\cite{prasad2000stereoscopic,wang2020stereo} and 3D Scanning PIV (SPIV)~\cite{brucker19973d,hori2004high} are two examples of techniques that have been developed to measure in-plane 3D velocity fields. Other approaches target the measurement of 3D velocity fields in the volume, including defocus PIV~\cite{pereira2000defocusing,pereira2002defocusing,yoon20063d}, synthetic aperture PIV~\cite{belden2010three}, tomographic PIV (tomo-PIV)~\cite{elsinga2006tomographic,scarano2012tomographic}, structured-light PIV~\cite{xiong2017rainbow,aguirre2019single,xiong2020rainbowpiv}, plenoptic (light field)  PIV~\cite{ding2023full,li20203d}, and Holographic PIV (HPIV)~\cite{trolinger1969holographic,brady2009compressive,mallery2019regularized,chen2020physics,chen2021holographic,chen2022compact,chen2023h}.
% In our application, we only demonstrate our approach in 2D using basic PIV technique, but we may extend to 3D applications, in the future.

\medskip\noindent
{\bfseries Imaging optical distortion.}
Imaging optical distortion is a critical task in several imaging
applications like microscopy, telescopes, and machine vision
systems. Optical distortion occurs when light rays are refracted
during their path, causing deviations from the ideal rectilinear light
propagation. Several techniques have been developed to correct optical
distortion since it can lead to inaccurate measurements, blurred
images, and reduced resolution. However, in some applications, optical
distortion is not a nuisance but rather a valuable tool. For example,
in techniques such as Background-Oriented Schlieren (BOS)
imaging~\cite{richard2001principle,meier2002computerized,vinnichenko2023performance},
wavefront sensing~\cite{wang2018megapixel,wang2019quantitative}, and
phase retrieval~\cite{candes2015phase,candes2015phasewf}, the
distortion is leveraged to reconstruct a signal of interest. One
solution to these problems is to capture the distortion of a patterned
background viewed through the transparent medium of interest (i.e.,
gas, optical system, etc.). By comparing images of the undisturbed and
disturbed background, these techniques can quantitatively measure the
displacements induced by the distortion and, therefore, infer the
underlying dynamics inside the medium.

\section{Method}

In the following we first describe the mathematical concept behind
Helmholtz decomposition, then introduce the HDNet architecture, and
finally introduce Helmholtz Synthesis as way of efficiently generating
training data. 

\begin{figure*}[htbp]
\centering\includegraphics[width=\linewidth]{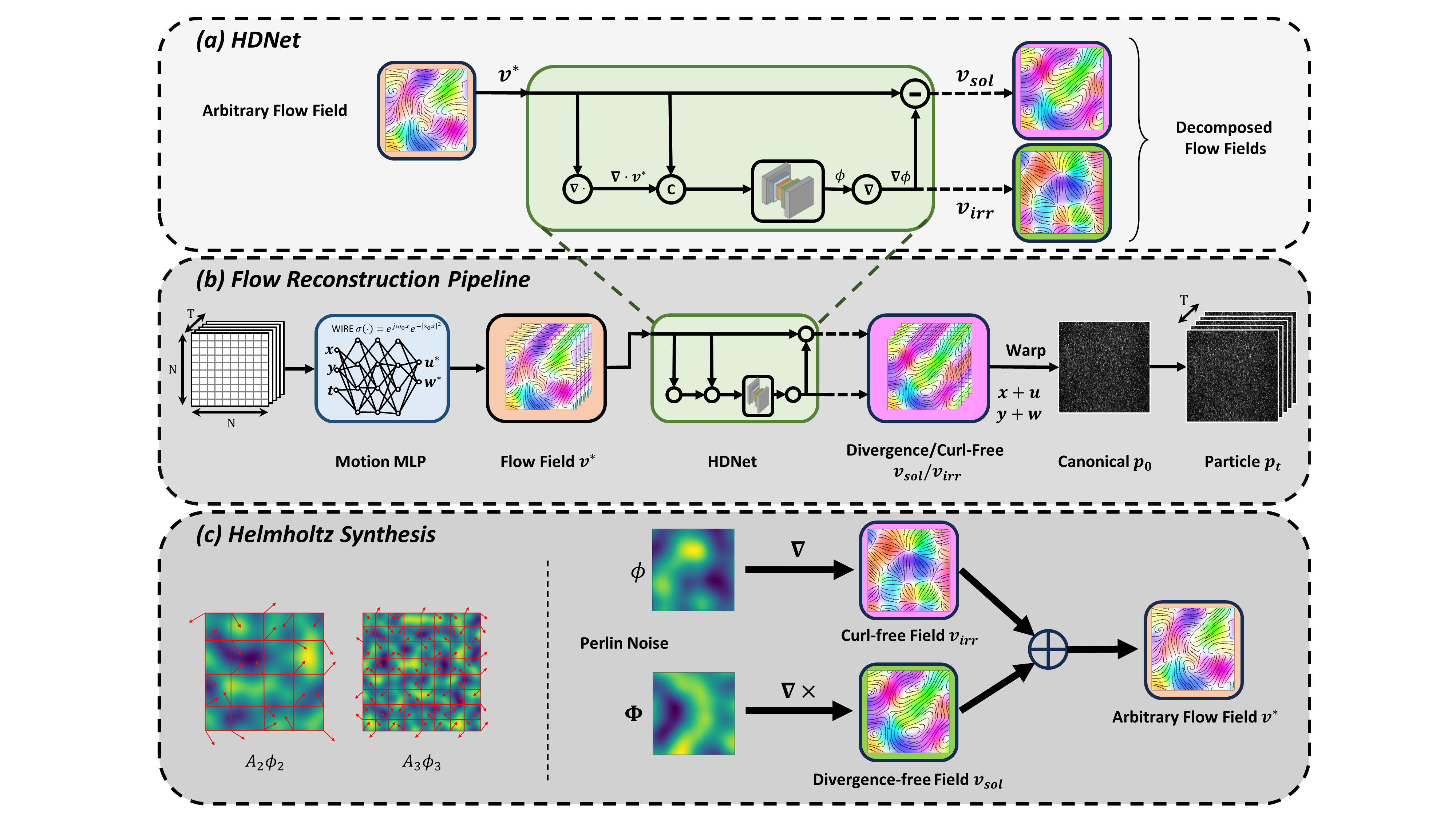}
\caption{Overview of our pipeline. (a) Concept of
  HDNet: an arbitrary flow field $\mathbf{ \VelocityField^*}$ is decomposed into
  its irrotational (curl free) and solenoidal (divergence free)
  components. This is known as a Helmholtz Decomposition (see text for
  details). (b) An example of how HDNet can be used to enforce
  the physical properties of vector fields in flow reconstruction
  problems. A motion MLP is trained to represent the flow of particles
in a fluid. The output of the MLP is not necessarily physically
plausible. Physical plausibility can be restored by utilizing HDNet to
extract the incompressible (i.e. divergence free) component of the
flow, which is then used to establish the temporal relationship
between the particle fields over time (see Section~\ref{PIV2Sec:Experiments}). (c)
HDNet is trained exclusively on synthetic data, generated by a reverse
process we term ``Helmholtz Synthesis'': irrotational and solenoidal
components are synthesized from spatially smooth random scalar fields
generated with Perlin Noise~\cite{bridson2007curl}. They can be combined into
general flows $\mathbf{ \VelocityField^*}$, forming training pairs with a known
Helmholtz decomposition, without requiring costly fluid simulation.}
\label{PIV2Fig:teaser}
\end{figure*}

\subsection{Helmholtz Decomposition and Physical Interpretation}\label{PIV2Sec:HDMath}
The key idea behind our approach is to utilize the Helmholtz
decomposition of vector fields, which is based on the
\emph{fundamental theorem of vector calculus}. This theorem states
that any arbitrary vector field $\vgen$ can be decomposed into two
orthogonal components: an \emph{irrotational} (curl-free) and a
\emph{solenoidal} (divergence-free) vector field:
\begin{equation}
  \begin{aligned}
    \vgen= \virr+\vsol.
\end{aligned}
\label{PIV2Eq:HZD}
\end{equation}
Classically, the Helmholtz decomposition is computed by leveraging well-known identities from vector calculus, and then (numerically) solving a Poisson equation. Specifically, any irrotational flow can be expressed as the gradient of a scalar \emph{potential field} $\potential$:
\begin{equation}
  \virr = \nabla\phi,
  \label{PIV2Eq:curlfree}
\end{equation}
Additionally, the curl of a gradient field is equal to zero, as is the divergence of a curl field:
\begin{equation}
  \curl(\nabla\potential)=0\quad\quad\mathrm{and}\quad\quad
  \divv(\curl\vv)=0.
  \label{PIV2Eq:Identities}
\end{equation}
By applying the divergence operator to both sides of Eq.~\ref{PIV2Eq:HZD}, we obtain:
\begin{equation}
  \divv\vgen = \divv\virr + \divv\vsol = \divv\virr,
\end{equation}
which yields the Poisson equation:
\begin{equation}
  \divv\vgen = \laplac\potential.
\label{PIV2Eq:Poisson}
\end{equation}
Once the potential field $\potential$ is retreived using a Poisson solver, the component fields can be calculated as follows:
\begin{equation}
  \virr=\nabla\potential \quad\quad\mathrm{and}\quad\quad
  \vsol=\vgen-\virr.
  \label{eq:esthd}
\end{equation}
It is important to note that the potential field $\potential$ has physical significance in many different application domains. For example in fluid simulation $\potential = P/\rho$ represents the normalized pressure, where $P$ is the pressure and $\rho$ is the mass density. This term plays a crucial role in the incompressible Navier-Stokes equations, which govern many physical fluid flows. In incompressible fluid simulation, a preliminary flow estimate $\vgen$ is often forced to be divergence-free (incompressible) by computing the $\vsol$ term according to Eq.~\ref {eq:esthd}, a process commonly known as the \emph{pressure projection} step.

In some optical applications, such as wavefront
sensing~\cite{wang2019quantitative}, phase
retrieval~\cite{candes2015phase}, or Background-Oriented Schlieren
imaging~\cite{richard2001principle}, the Helmholtz decomposition takes
on another physical interpretation. In these applications, the flow
fields correspond to \emph{optical flow}, which describes how light
rays bend due to an optical distortion. This distortion is caused by a
spatially varying phase delay in the optical wavefront. The potential
field $\phi$ is precisely this phase profile, while the observed
optical flow is proportional to its gradient, $\nabla\phi$, making it
an irrotational field (see Supplement for details). Therefore, when
reconstructing optical flow for an optical distortion inverse problem,
it is valid to employ the Helmholtz decomposition to ensure the
reconstructed flow is curl-free.

However, the classical Poisson solver approach is not differentiable, making it challenging to integrate into modern PINN pipelines for either forward simulation or inverse reconstruction tasks. 
With the recent advancements in deep learning, numerous deep learning solvers for PDEs have been proposed~\cite{sirignano2018dgm,raissi2019physics,caiphysics,tompson2017accelerating,cai2021flow,cai2021artificial}. Inspired by these ideas, we propose HDNet (Helmholtz decomposition Network) a novel neural network designed to perform Helmholtz decomposition based on conventional HD (Helmholtz decomposition) operations as described by Eqs.\ref{PIV2Eq:Poisson} and \ref{eq:esthd}.

\subsection{HDNet}
\paragraph{Architecture.}
The HDNet architecture, depicted in Fig.~\ref{PIV2Fig:teaser} (a), takes as input either an arbitrary flow field or an "initial" estimation of the reconstructed flow. Instead of relying on the commonly used iterative Poisson solver, HDNet employs a deep learning (DL) solver based on a Convolutional Neural Network (CNN) encoder-decoder with a UNet architecture. To mitigate grid artifacts caused by max-pooling layers, we replace them with convolutions with stride, and similarly, we replace all the transpose convolutions with up-sampling layers.

From Equation~\ref{PIV2Eq:Poisson}, the input of the network is a
general velocity field $\vgen$. To facilitate learning of the
Helmholtz decomposition, we compute the divergence $\divv\vgen$, and
concatenate it with this input. The UNet core of the HDNet takes this
concatenated input and computes the desired potential field
$\potential$.  From Equations~\ref{PIV2Eq:HZD}
and~\ref{PIV2Eq:Identities}, we can compute the gradient of the output
scalar $\nabla\potential$ to get the curl-free component $\virr$. By
subtracting this curl-free component from the input field, we obtain
the divergence-free component ($\vsol$).

\paragraph{Training loss.}
To train the proposed network, we aim to minimize the discrepancy between the predicted scalar field $\hat{\ScalarPotential}$ and the ground truth scalar field $\ScalarPotential$, while simultaneously minimizing the difference between the predicted divergence-free field component ($\mathbf{\hat{\VelocityField}_{sol}}$) and the ground truth component ($\mathbf{\VelocityField_{sol}}$). Thus, the training loss is defined as follows:
\begin{equation}
\begin{aligned}
\mathcal{L}= \lVert \mathbf{\hat{\VelocityField}_{sol}} - \mathbf{\VelocityField_{sol}} \rVert_2^2+ \lossweight \lVert \hat{\ScalarPotential} - \ScalarPotential \rVert_2^2.
\end{aligned}
\end{equation}
The ground truth $\mathbf{\VelocityField_{sol}}$ and $\ScalarPotential$ come from synthesized training data which will be introduced in the following subsection.

\subsection{Training Data Generation with Helmholtz Synthesis}
To enable supervised training of the network, it is necessary to generate training pairs for HDNet. The training result highly relies on the training dataset quantity and quality. However, conventional commercial fluid simulation software are computationally expensive and time-consuming, limiting the generation of large-scale datasets. To overcome this challenge, we introduce the \emph{Helmholtz synthesis} module, a novel approach for generating large-scale fluid training datasets. The Helmholtz synthesis module generates data through the reverse process of the Helmholtz decomposition. By exploiting the vector identities in Eq.~\ref{PIV2Eq:Identities}, we can compute irrotational and solenoidal fields from pseudo-random scalar fields and combine them to obtain arbitrary flow fields. To ensure a realistic quality of the generated dataset, we require scalar fields that are smooth, bandwidth-limited, and exhibit large variability, thereby mimicking real-world complexity. 

\begin{figure}[htp]
\centering\includegraphics[width=\linewidth]{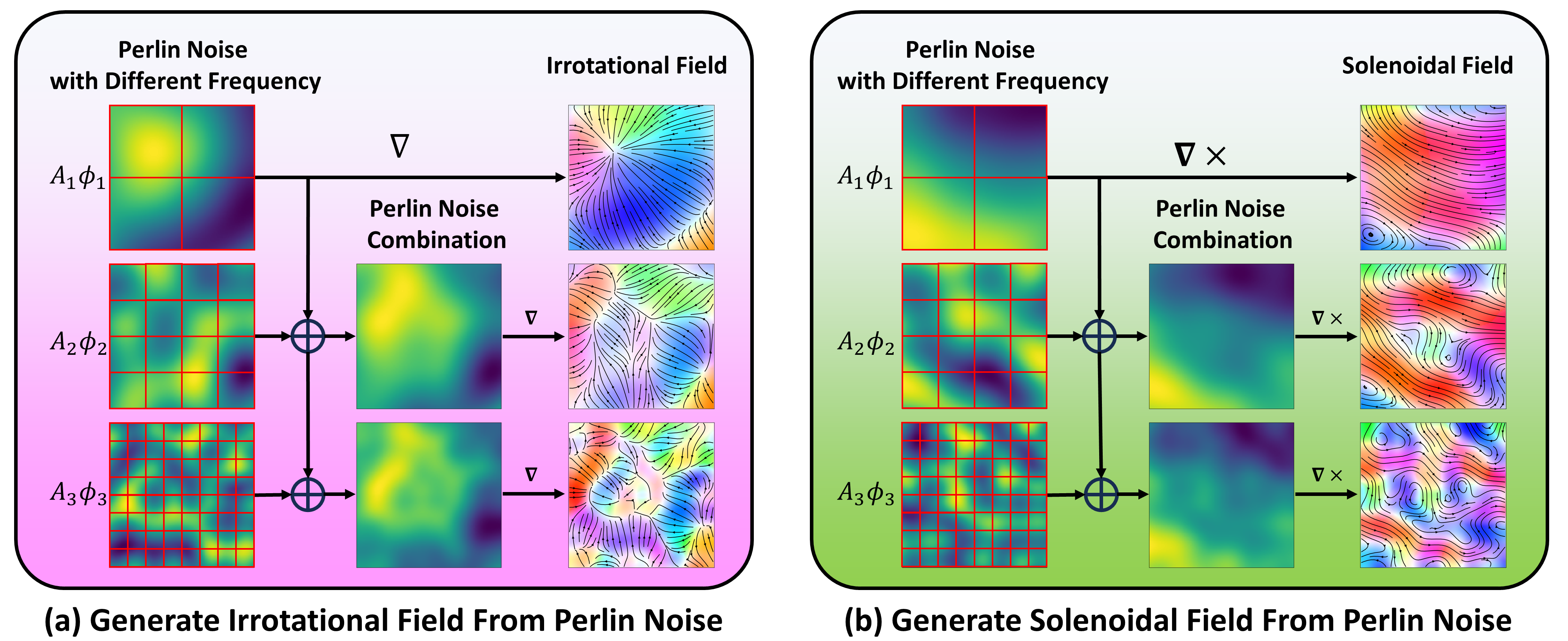}
\caption[Perlin noise]{Perlin noise. The red grid in the Perlin noise subfigures controls the Perlin noise frequency. $\ScalarPotential_1$ means with grid $2 \times 2$. More grid number means high frequency Perlin noise. (a) is Perlin noise for generating irrotational field. $\mathbf{\nabla}$ mean the gradient of Perlin noise. (b) is for solenoidal field $\mathbf{\nabla \times}$ mean the curl of Perlin noise.}
\label{PIV2Fig:PerlinNoise}
\end{figure}

\paragraph{Perlin Noise.} In Computer Graphics, Perlin noise~\cite{perlin1985image}
was developed as a way to produce random band-limited scalar fields
that satisfy our requirements for spatial smoothness. Different
frequency bands of Perlin noise can be composited to generate random
patterns of varying spatial detail, known as ``turbulence'' in the
graphics literature.  Thus, the pseudo-random generated
scalar fields are expressed as: $\ScalarPotential=\sum
A_{n}\ScalarPotential_{n}$, where $A_{n}=1/{2^n}$ is the amplitude of
the Perlin noise for the scale $n$, and $\ScalarPotential_{n}$ is the
Perlin noise generated with a $2^n \times 2^n$ grid. The grid number
$2^n$ controls the frequency of the generated Perlin noise: the higher
the grid number, the higher the frequency of the Perlin noise, and the
smaller its amplitude. This amplitude rule is designed in such a way
to better simulate natural signals. For example, in
Fig.\ref{PIV2Fig:PerlinNoise}, $A_2\ScalarPotential_2$ denotes a
Perlin noise with a grid number of $4$ and an amplitude of
$1/4$. While $A_3\ScalarPotential_3$ denotes a Perlin noise with a
grid number of $8$ and an amplitude of $1/8$. 

In our application, we require a Perlin turbulence scalar field $\ScalarPotential$ to generate the irrotational velocity field. We also need a Perlin turbulence vector field $\mathbf{\VectorPotential}=(\VectorPotential_1,\VectorPotential_2,\VectorPotential_3)$, composed of three Perlin turbulence scalar fields $\VectorPotential_1,\VectorPotential_2,\VectorPotential_3$, for the generation of the solenoidal velocity field. Therefore, the irrotational and the solenoidal velocity fields are respectively constructed as the gradient of the Perlin noise scalar field~\cite{bridson2007curl}, and the curl of the Perlin noise vector field:
\begin{equation}
\begin{aligned}
\virr=\nabla\ScalarPotential \quad\quad\mathrm{and}\quad\quad
  \vsol = \nabla \times \mathbf{\VectorPotential}.
\end{aligned}
\end{equation}
% We can generate a Perlin noise $\ScalarPotential$ for irrotational field. Generate another three Perlin noises $\VectorPotential_1,\VectorPotential_2,\VectorPotential_3$ which are composited as a vector field $\mathbf{\VectorPotential}=(\VectorPotential_1,\VectorPotential_2,\VectorPotential_3)$ for solenoidal field.
% \writingidea{ This method enables the efficient generation of large-scale paired datasets, which enable the supervised learning of the network.}
% We construct a solenoidal velocity field as the curl of the Perlin noise vector field:
% \begin{equation}
% \begin{aligned}
% \mathbf{\VelocityField_{sol}}=\nabla \times \mathbf{\VectorPotential}.
% \end{aligned}
% \end{equation}
Note that for 2D flows, $\mathbf{\VectorPotential}$ degenerates into a scalar field $\VectorPotential$, and the solenoidal velocity field is constructed as:
\begin{equation}
\begin{aligned}
\mathbf{\VelocityField_{sol}}=(\frac{ \partial \VectorPotential}{\partial \yaxis}, -\frac{ \partial \VectorPotential}{\partial \xaxis}).
\end{aligned}
\end{equation}
% The irrotational velocity field is constructed as the gradient of the Perlin noise scalar field:\writingidea{Maybe say is defined as Expression~\ref{PIV2Eq:curlfree} to avoid repeat}
% \begin{equation}
% \begin{aligned}
% \mathbf{\VelocityField_{irr}}=\nabla \ScalarPotential.
% \end{aligned}
% \end{equation}
From the vector identities in Eq.~\ref{PIV2Eq:Identities}, we know that the constructed fields satisfy the following properties: $\mathbf{\VelocityField_{sol}}=\nabla \times \mathbf{\VectorPotential}$ is divergence-free (divergence of curl is zero) and $\mathbf{\VelocityField_{irr}}=\nabla \ScalarPotential$ is curl-free (curl of gradient is zero). By combining these two fields, we can obtain an arbitrary flow field using Helmholtz decomposition:
\begin{equation}
\begin{aligned}
\mathbf{\VelocityField^*}=\mathbf{\VelocityField_{irr}}+\chi \cdot \mathbf{\VelocityField_{sol}},
\end{aligned}
\end{equation}
where $\chi$ is the weight controlling the relative strengths of the
two components.

During the training, we use this generated flow field $\mathbf{\VelocityField^*}$ as the input for our HDNet, while the corresponding divergence-free field $\mathbf{\VelocityField_{sol}}$ and the scalar field $\ScalarPotential$ are used as the ground truth of the network. 
Using this method, we can generate the $20000$ fluid training data pairs with a resolution of $128 \times 128$ within approximately half an hour.

\section{Application: Use of HDNet for Flow Reconstruction}
The proposed HDNet network is versatile and can be incorporated into
any differentiable flow simulation or reconstruction pipeline, to
enforce hard constraints on the physical properties of the flow in
various applications. In this work we are primarily interested in
inverse problems that can be expressed as flow estimation tasks. We
show quite different applications from fluid flow to optical
distortion imaging, all using the same general experimental framework
outlined in the following. 

%In this section, we propose an example of a differentiable flow reconstruction pipeline incorporating HDNet. We demonstrate its effectiveness in solving some 2D flow estimation problems, namely PIV and BOS. \RI{Add phase estimation, if included in the experiment part.}

Our pipeline consists of a Physics-Informed Neural Network (PINN) for
flow reconstruction, as illustrated in Fig.~\ref{PIV2Fig:teaser}
(b). First, we use a coordinate-based MLP network to represent the
flow field $\mathbf{\VelocityField^*} = \MLP(\xaxis, \yaxis, \taxis;
\nnweight)$, where $\nnweight$ is the Motion MLP network weight. This
network takes as input the spatial and temporal coordinates $(\xaxis,
\yaxis, \taxis)$, and outputs an "initial" reconstructed motion field
$\mathbf{\VelocityField^*} = (\Velocityfirstcomponent^*,
\Velocitysecondcomponent^*)$ for each frame.

We employ the Wavelet Implicit neural REpresentation (WIRE)~\cite{saragadam2023wire} for the Motion MLP, which utilizes a Gabor wavelet as the activation function to learn high-frequency flow motion. Indeed, this activation function has a controllable parameter $\frepara$ that represents the frequency of the signal.
% \paragraph{Coarse-to-fine Strategy.} 
By adjusting $\frepara$ during the learning process, we can achieve a coarse-to-fine reconstruction. A smaller $\frepara$ generates smoother results, corresponding to coarse reconstruction, while a larger $\frepara$ generates high-frequency details, corresponding to fine reconstruction.
More details about the WIRE representation and the coarse-to-fine reconstruction strategy are respectively discussed in the Supplement Sections~\ref{PIV2Sec:coarse-to-fine}~and~\ref{PIV2Sec:WIRE}.
% A smaller $\frepara$ generates smoother results corresponding to the ``coarse" reconstruction. A large $\frepara$ generates more high-frequency detail corresponding to ``fine" reconstruction. By using this property, we can achieve coarse-to-fine reconstruction. 
% More details are discussed in the Supplement Sections~\ref{PIV2Sec:WIRE~and~\ref{PIV2Sec:coarse-to-fine}.

In the applications we investigate, the reconstructed flows are not
arbitrary but exhibit specific physical properties. For example, in
the particle imaging velocimetry (PIV) application, the flow of
incompressible fluids is divergence-free, while in phase distortion problem imaging, the gradient of the air refractive index
reconstruction is curl-free (see Section \ref{PIV2Sec:HDMath}). To
impose these physical constraints on the initial reconstruction, we
apply the pre-trained HDNet to the velocity field
$\mathbf{\VelocityField^*}$. The output of the HDNet is the pair of
the irrotational and solenoidal fields:
$\ProjHDNet(\mathbf{\VelocityField^*}) =
(\mathbf{\VelocityField_{sol}},
\mathbf{\VelocityField_{irr}})$. According to the application, we
select the component of the velocity that satisfies the physical
constraints.  Using the HDNet output
($\mathbf{\VelocityField_{sol}}/\mathbf{\VelocityField_{irr}}$), we
can warp the canonical template field $\ReflectiveIndex_0$ to obtain
the scene field $\ReflectiveIndex_{\taxis}$ for each
frame. Mathematically, this process is expressed as:
% \begin{equation}
% \mathbf{\ReflectiveIndex_t(x,y)}=\mathbf{\ReflectiveIndex_0 (x_0+\Velocityfirstcomponent^t_{sol/irr},y_0+\Velocitysecondcomponent^t_{sol/irr})},
% \label{PIV2Eq:NonLinBC}
% \end{equation}
% \writingidea{If not enough space I can just describe by words for zeros.}
% where 
% \[ \mathbf{\VelocityField^t_{sol/irr}}= \begin{cases} 
%       \mathbf{0}    &\text{if $t=0$}\\
%       (\mathbf{\Velocityfirstcomponent^t_{sol/irr}}, \mathbf{\Velocitysecondcomponent^t_{sol/irr}})   &\text{if $t \neq 0$}.
%    \end{cases}
% \]
\begin{equation}
\hat{\ReflectiveIndex}_{\taxis}(\rVector)=\ReflectiveIndex_0 (\rVector+\mathbf{\VelocityField_{sol/irr}}),
\label{PIV2Eq:NonLinBC}
\end{equation}
where $\rVector=(\xaxis, \yaxis)$ is the spatial
coordinates. $\mathbf{\VelocityField}^{\taxis}_{\mathbf{sol/irr}}$ is
the HDNet output: $\mathbf{\VelocityField_{sol}}$ or
$\mathbf{\VelocityField_{irr}}$ at time
$\taxis$. $\hat{\ReflectiveIndex}_{\taxis}(\rVector)$ is the predicted
``scene'' field at time $\taxis$. It represents the particle density
field in the PIV application or the background texture in the case of
BOS or wavefront sensor problem. $\ReflectiveIndex_0$ is the canonical configuration of the
``scene'' field. In our framework, we represent this reference
configuration using simply a template image, which is a variable to be
optimized during the learning process. Eq.~\ref{PIV2Eq:NonLinBC}
is equivalent to the non-linearized brightness
consistency~\cite{brox2004high} in the optical flow
problems. Therefore, our pipeline inherently incorporates the
non-linearized brightness consistency.  
Our pipeline is defined as a joint optimization problem, where we aim to retrieve both the canonical scene field and the motion field representation:
% \begin{equation}
% \begin{aligned}
% \mathcal{L}= \lVert \mathbf{\hat{\ReflectiveIndex}_0(x_0+\Velocityfirstcomponent^t_{sol/irr},y_0+\Velocitysecondcomponent^t_{sol/irr})- \ReflectiveIndex_t}  \rVert_2^2  + \lossgradweight \mathbf{\lVert \nabla_{x,y} \VelocityField_t \rVert } 
% \end{aligned}
% \end{equation}
% \writingidea{Maybe change the above to be argmin problem and come from the implicit neural representation (weight w) to the end.}
\begin{equation}
\begin{aligned}
\mathcal{L}= \argmin_{\ReflectiveIndex_0,\nnweight} \sum_{\xaxis, \yaxis, \taxis} \lVert \ReflectiveIndex_0(\mathbf{r}+\ProjHDNet(f(\xaxis, \yaxis, \taxis; \nnweight)))- \ReflectiveIndex_{\taxis} \rVert_2^2 + \lossgradweight \lVert \nabla_{\xaxis, \yaxis} \ProjHDNet(f(\xaxis, \yaxis, \taxis; \nnweight)) \rVert 
\label{PIV2Fig:pipelineloss}
\end{aligned}
\end{equation}
The second term corresponds to the total variation (TV) of the velocity field, which promotes smoothness in the velocity field, limiting changes within the neighborhood. $\lossgradweight$ is the weight of the TV term in the total loss.

\section{Experiments}\label{PIV2Sec:Experiments}
In this section, we demonstrate the application of HDNet within our flow reconstruction pipeline. We first evaluate the reconstruction performance HDNet performance and flow reconstruction pipeline of by using synthetic data in order to assess numerical metrics. Then, we showcase reconstruction outcomes obtained from real PIV, and BOS experimental data, providing visualizations of divergence and curl maps alongside with scalar potential fields.
Similar experiments are also presented in the Supplement (Section~\ref{PIV2Fig:PhaseDemo}) for another phase distortion problem: wavefront sensing.
The obtained results demonstrate the superior performance of our method in terms of reconstruction quality, physical constraint enforcement, and robustness against noise.
% In this section, we will take the PIV and the BOS imaging as examples to illustrate how to use HDNet in our pipeline for flow reconstruction while preserving the physical constraint. We will first show the reconstruction comparison of different method and setting by synthetic PIV data, accompanied by numerical metric evaluations. Then, we showcase the reconstruction outcomes with real PIV and BOS experiment data, providing divergence and curl maps, along with scalar fields. The results show that our method has better performance in reconstruction quality, enforcing physical constraint, and robustness against noise. These experiments show our method's versatility and flexibility in satisfying both curl-free and divergence-free constraints, while also outputting the scalar field simultaneously.

% \subsection{Helmholtz Synthesis}

% \writingidea{Show the figure of Helmholtz Synthesis result of different setting. Like different Perlin noise frequency, different divergence weight}

\subsection{Synthetic Data}
\begin{table}[h]
  \caption{MSE value of HDNet inference divergence}
  \label{PIV2Table:HDNetEva}
\centering
\resizebox{\columnwidth}{!}{%
\begin{tabular}{@{}llllll@{}}
\toprule
          & $n=2,\chi=10^{-5}$   & $n=3,\chi=10^{-5}$    & $n=4,\chi=10^{-5}$    & $n=4,\chi=10^{-4}$   & $n=4,\chi=10^{-3}$    \\ \midrule
Mean      & $1.4 \times 10^{-7}$ & $3.87 \times 10^{-7}$ & $5.71 \times 10^{-7}$ & $9.31\times 10^{-7}$ & $1.24 \times 10^{-6}$ \\
Std. Dev. & $6.8 \times 10^{-8}$ & $7.2 \times 10^{-8}$  & $4.1 \times 10^{-8}$  & $1.27\times 10^{-7}$ & $2.36 \times 10^{-7}$ \\ \bottomrule
\end{tabular}
}
\end{table}
\medskip\noindent
{\bfseries HDNet evaluation.}
We demonstrate the performance of HDNet in enforcing incompressibility on the outputed solenoidal field using synthetic data generated by Helmholtz synthesis module. We create five groups of paired datasets, each containing 800 samples, with varying Perlin noise scales $n$ and the weight $\chi$. We assess the divergence of the solenoidal field $\vsol$ output by HDNet and report the mean squared error (MSE) with respect to zero (mean and standard deviation) in Tab.~\ref{PIV2Table:HDNetEva}.
% To have a quantitative evaluation of our HDNet physical constraint performance, we generate the evaluation dataset by Helmholtz synthesis. 5 groups of different Perlin noise scales $n$ and $\chi$ are generated. Each group contains 800 data. The input field is a composited arbitrary field (with divergence and curl value). We evaluate the divergence of HDNet output solenoidal field $\vsol$. The divergence MSE (mean square error) value with zero is given in the Table~\ref{PIV2Table:HDNetEva}. 

\medskip\noindent
{\bfseries Flow pipeline evaluation.}
To quantitatively assess the performance of our flow reconstruction pipeline, we generated synthetic PIV particle image sequences as described in the Supplement~\ref{PIV2Sec:PIVSynPar}. We then compare the predicted flows using different reconstruction methods with the ground truth one.
% To have some quantity assessment of our pipeline performance, we generated some PIV particle image sequence (see detail in Supplement~\ref{PIV2Sec:PIVSynPar}). Then, these particle image sequence are ground truth $\ReflectiveIndex$ in Equation~\ref{PIV2Fig:pipelineloss} that our predicted particle image sequence to compare with.
\begin{figure}[htp]
\centering\includegraphics[width=\linewidth]{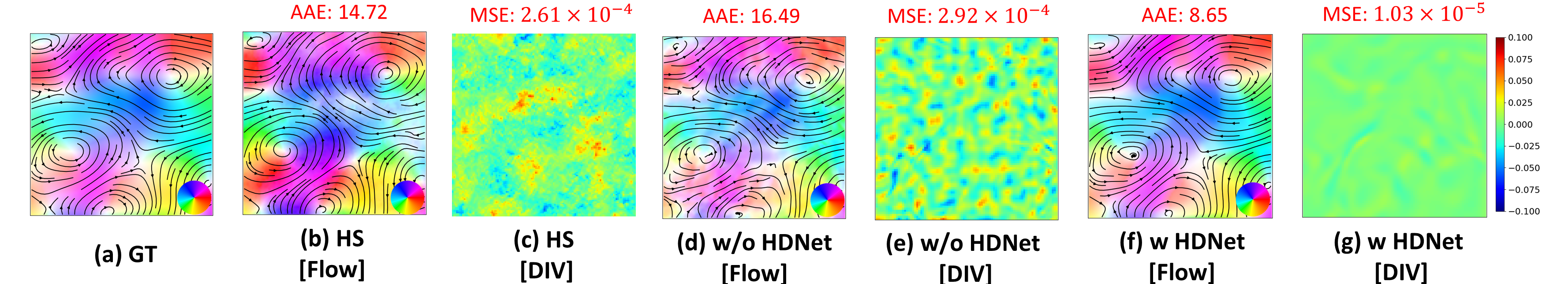}
\caption[Synthetic Data Reconstruction]{Synthetic PIV data reconstruction comparison. For each method we illustrate the flow field (Flow) and its divergence (DIV). (a) Ground Truth, (b-c) Horn-Schunck optical flow reconstruction, (d-e) our pipeline without HDNet, (f-g) our pipeline with HDNet. We also include quantitative evaluations: AAE (average angular error), and MSE (mean squared error of the divergence).}
\label{PIV2Fig:SyntheticPIV}
\end{figure}
% [Old caption] Synthetic PIV data reconstruction comparison. w/o HDNet means our flow reconstruction pipeline without HDNet. w HDNet mean our flow reconstruction pipeline with HDNet. HS means Horn-schunck optical flow reconstruction. DIV mean the divergence of left velocity field figure. AAE is the average angular error. MSE is the mean square error of the divergence.
The results of this comparison are reported in
Fig.\ref{PIV2Fig:SyntheticPIV}. When using our pipeline without HDNet,
the reconstruction exhibits significant artifacts and has larger
errors than the baseline flow reconstruction using
Horn-Schunck~\cite{horn1980determining} optical flow, in both used
metrics: AAE (Average Angular Error value) for the flow and the MSE
(Mean Square Error) for the divergence. When using the HDNet, our
pipeline improves the flow reconstruction with a better AAE, while
at the same time reducing the divergence MSE by more than an order of
magnitude.

% Fig.\ref{PIV2Fig:SyntheticPIV} is the reconstruction comparison. w/o HDNet means our flow reconstruction pipeline without HDNet. We can see the reconstruction have a lot of artifacts. DIV mean the divergence of left subfigure. AAE is the average angular error value~\cite{chen2021snapshot}. It is a common metric to evaluate the flow reconstruction. MSE is the mean square error of the divergence. We can see from (b) without HDNet, the divergence is still large. HS mean Horn–Schunck~\cite{horn1980determining} optical flow reconstruction. From (c), we can see that the reconstruction still have some artifacts, especially not that continuous. The HS divergence is still large. w HDNet means our flow reconstruction pipeline with HDNet. We can see reconstruction is very close to the ground truth. The divergence is also very small. This outcome demonstrates that the integration of HDNet significantly enhances the performance of our flow reconstruction and enforce reconstruction to satisfy the physical constraints.

\subsection{Real Experiment Data}
We also verify our flow reconstruction pipeline in real experiments
for both PIV and optical distortion applications. Details of the
experimental settings and image sequences can be found in
Supplement Section~\ref{PIV2Sec:Imple} and
Fig.~\ref{PIV2Fig:RealParticle}.

\subsubsection{PIV}
\begin{figure}[htp]
\centering\includegraphics[width=\linewidth]{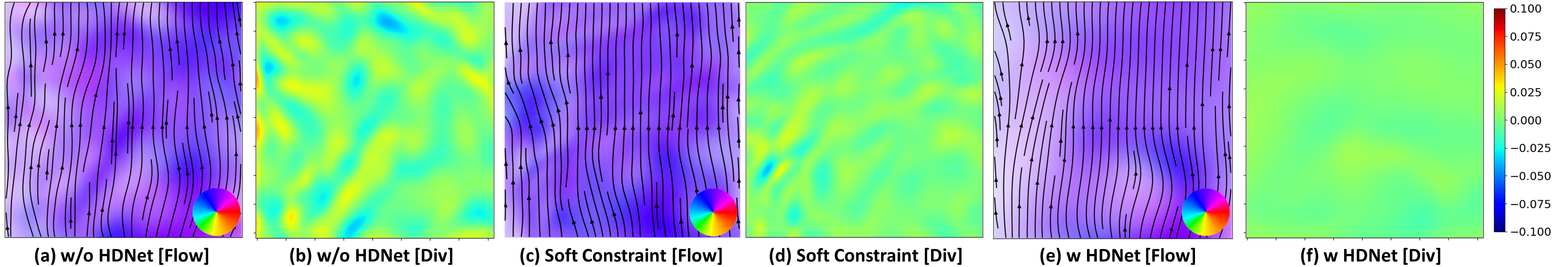}
\caption[Real PIV experiment]{Real PIV data reconstruction comparison. Soft constraint is the method that add a divergence penalty term to the total loss.}
\label{PIV2Fig:RealPIV}
\end{figure}

Fig.~\ref{PIV2Fig:RealPIV} shows the real PIV experiment flow
reconstructions. Here, ``Soft Constraint'' reconstruction consists of adding
an $\ell_2$ norm of the divergence ($\mathbf{\lVert \nabla \cdot
  \VelocityField_t } \rVert_2^2$) to the total loss
(Equation~\ref{PIV2Fig:pipelineloss}) to penalize the divergence. It
is a very straightforward idea, but the solution does not always
satisfy the constraint. From Fig.~\ref{PIV2Fig:RealPIV} (d), we can
see that there is still some divergence error. With the help of HDNet,
our pipeline can reconstruct the flow very well and remove the
diverging error. Our method can be thought of as a differentiable hard
constraint for the flow reconstruction.

\subsubsection{Background-Oriented Schlieren Imaging}
\begin{figure}[htp]
\centering\includegraphics[width=\linewidth]{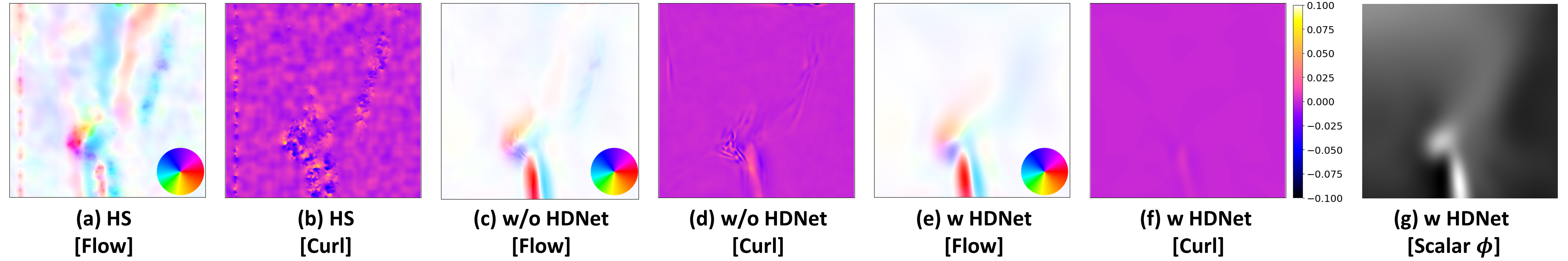}
\caption[BOS Reconstruction]{BOS reconstruction comparison. Curl mean
  the curl value map of left figure. w HDNet scalar $\ScalarPotential$
  means our flow reconstruction pipeline with HDNet and the scalar
  output of HDNet, that clearly shows the hot air plume above the candle. }
\label{PIV2Fig:SchilierenImaging}
\end{figure}

In optical distortion applications like BOS, the reconstructed optical flow is the gradient of refractive index
(phase)~\cite{atcheson2008time}, also see
Supplement~\ref{PIV2Sec:PhasePrin}. Therefore, this reconstructed flow should
be curl-free (i.e., the curl of the gradient is zero). We compare
different methods of reconstruction in
Fig.\ref{PIV2Fig:SchilierenImaging}, and illustrate the reconstructed flows and their corresponding curls .

The flow reconstruction pipeline is similar to the one used for PIV experiments. The only differences are the input images (see Fig.~\ref{PIV2Fig:BOSimgs} in Supplement), and the use of the $\virr$ output of HDNet instead of the $\vsol$ output.
The compressed video data used in the original paper~\cite{atcheson2008time} exhibits some compression artifacts, leading to noisy flow reconstruction using traditional methods like Horn-Schunck (Fig.~\ref{PIV2Fig:SchilierenImaging} (a,b)). Our proposed PINN flow reconstruction pipeline produces a significantly cleaner flow field, effectively removing even the vertical line artifact present in the HS reconstruction Fig.~\ref{PIV2Fig:SchilierenImaging} (a-b). Without the use of HDNet, the PINN flow reconstruction pipeline still exhibits a relatively large curl error (see Fig.~\ref{PIV2Fig:SchilierenImaging} (d)). By incorporating HDNet, we achieve a significantly smaller curl error Fig.~\ref{PIV2Fig:SchilierenImaging} (f), indicating a highly accurate and physically consistent flow reconstruction.
Notably, by using HDNet, our method not only reconstructs the flow but also recovers the corresponding phase as illustrated in Fig.~\ref{PIV2Fig:SchilierenImaging} (g).
% Because the data was extracted from compressed video data in BOS paper~\cite{atcheson2008time}, it suffered from compressed artifacts. From Fig.~\ref{PIV2Fig:SchilierenImaging} (a), we can see that the Horn–Schunck optical flow reconstruction is very noisy, resulting in a large curl error (b). But our method (e) is very clean. There is a vertical line artifacts in the left part of (a), which comes from  structure wavelet pattern (see Supplement Fig.~\ref{PIV2Fig:BOSimgs}). Even this artifacts is removed in our method (e). From (d), without using the HDNet, the PINN flow reconstruction pipeline still has a relatively large curl error. Comparing (c) and (e), there is some artifacts in (c). From (f), we can see that by using HDNet, the flow reconstruction curl error is very small which means that we have good reconstruction performance and satisfy the physical constraints.  (g) is the scalar field which is the phase in this example (see Section~\ref{PIV2Sec:HDMath}). By using HDNet, we can not only reconstruct the flow but also reconstruct the corresponding phase in this example. 
\section{Limitations and Future Work}

Although HDNet provides a convenient and effective way to inject
physical priors into PINNs, the current work has several limitations.
First, while the mathematical derivation of the approach holds both in
2D and 3D, currently only the 2D version is implemented. However, the
network architecture should be straightforward to adapt to 3D, and
since Perlin noise is also defined in 3D, data generation with
multi-scale Helmholtz Synthesis is also straightforward. Therefore we
do not expect that the generalization of HDNet to 3D will require more
than hyperparameter tuning.

A more fundamental limitation is that, as a supervised method, HDNet
does not guarantee an exact Helmholtz decomposition of the input flow;
in particular the solenoidal component is not guaranteed to be
strictly divergence free. The irrotational component is computed as
the gradient of an estimated potential field
($\mathbf{\VelocityField_{irr}}=\nabla \ScalarPotential$), and is
therefore always curl free. However, any mis-estimation of the
potential field $\ScalarPotential$ results in an imprecise
decomposition, and thus the calculated solenoidal flow
$\mathbf{\VelocityField_{sol}}=\mathbf{\VelocityField}^*-\mathbf{\VelocityField_{irr}}$
may still have a remaining divergence component. Our experiments show
that this effect is small, however if it is a concern in a particular
application, it is also possible to penalize
$\nabla\cdot\mathbf{\VelocityField_{sol}}$ in the loss function for a
larger PINN architecture.

\section{Conclusion}

In this paper, we propose HDNet, a novel network based on the \emph{fundamental theorem of vector calculus} and \emph{Helmholtz decomposition theorem}. By employing HDNet, we can effectively impose differentiable hard constraints on inverse imaging problem. We further propose the Helmholtz synthesis module that efficiently generates paired data by reversing Helmholtz decomposition. This module enables the rapid creation of 20000 data pairs within half an hour, making large-scale flow dataset construction feasible and the supervised training of HDNet possible.

Finally, we demonstrate the integration of HDNet into a PINN pipeline for flow reconstruction, showcasing its applicability with examples from PIV and BOS imaging data. Experimental results prove that our HDNet-empowered PINN pipeline outperforms conventional flow reconstruction method. 
Notably, our approach exhibits versatility and flexibility in satisfying both curl-free and divergence-free constraints while also outputting the scalar potential field.

\begin{ack}
The authors would like to thank Congli Wang, Ivo Ihrke and Abdullah Alhareth for providing data. This work was supported by KAUST individual baseline funding.
\end{ack}

\bibliographystyle{plain}
\bibliography{references}

%%%%%%%%%%%%%%%%%%%%%%%%%%%%%%%%%%%%%%%%%%%%%%%%%%%%%%%%%%%%

\appendix

\section{Implementation Details}\label{PIV2Sec:Imple}

The HDNet architecture is an MLP with 6 layers, 4 of which are hidden
layers. Each hidden layer has 64 neurons. For the WIRE activation
functions, the $\omega_0$ value ranges from 1.0 to 1.5. $\sigma_0$ value ranges from 0.8 to
1.2.  We choose the Adam optimizer with
learning rate is $1 \times 10^{-6}$. We train the HDNet on 20000 data
pairs, with 2000 data pairs as the evaluation dataset. Training takes
72 hours on a single A100 GPU. HDNet is applied after 30000 epoch after
the coarse reconstruction is almost done.

For the data generation, we use Perlin noise at scales $n$ from 1 to
5. The relative strength of the irrotational and the solenoidal,
controlled by the weight $\chi$, can be tuned to the specific
application. For example in PIV, the basic Horn Schunck optical flow
for an incompressible flow would already have a small divergence that
just needs to be reduced further. Therefore we can estimate
appropriate weights for the two terms by analyzing the divergence of
the basic flow estimate for a representative flow, and choose $\chi$
appropriately. Using this approach, we chose $\chi$ to be a random
number from a normal distribution with a mean of 0.0002.

For the full flow estimation pipeline, we chose one of the input
frames $\ReflectiveIndex_0$ as the template, and initialize
accordingly.

\section{Experiments}

\subsection{Phase Distortion Problem Principle}\label{PIV2Sec:PhasePrin}

\begin{figure}[htbp]
\centering\includegraphics[width=0.5\linewidth]{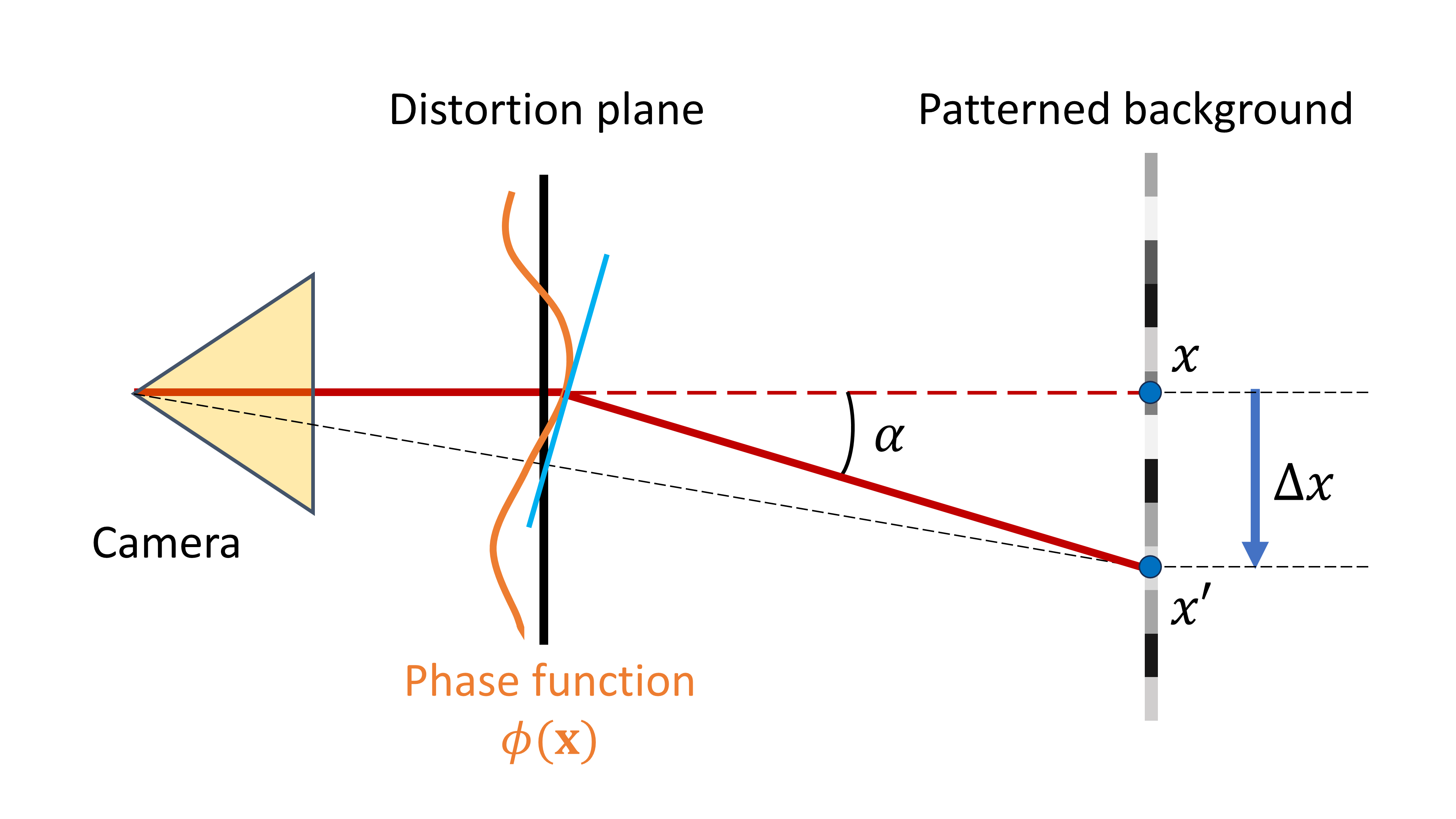}
\caption[BOS]{Distortion Imaging Workflow. The camera captures a video sequence of a patterned background through a distortion plane represented by the phase function $\phi(\mathbf{x})$. The patterned background at position $x'$ appears distorted to position $x$. Optical flow analysis of the captured video sequence calculates the displacement $\Delta x$, quantifying the distortion. The deflection angle $\alpha$ represents the deviation caused by the distortion plane, and it is proportional to the gradient of the phase $\phi$.\label{fig:bos}}

\subsection{Wavefront Sensor}

\label{PIV2Fig:PhaseDemo}

\end{figure}

Optical distortion imaging like BOS, wavefront sensing, and phase
retrieval can be approached in different ways, but one common approach
is to track the apparent motion (optical flow) of a high frequency
pattern imaged through the distortion. An example geometry for BOS is
shown in Figure~\ref{fig:bos}. The goal of BOS imaging is to measure
the phase delay in the distortion plane. A patterned background some
distance away is observed with a camera. Since light rays propagate
perpendicular to the phase profile $\potential$, the observed optical flow
is proportional to the $\nabla\potential$ for small angles (``paraxial
approximation''). The factor of proportionality is the propagation
distance between the distortion plane and the background.

Because the optical flow is proportional to the phase, the flow is
curl free, and the phase actually corresponds to the potential
function in the Helmholtz decomposition

\begin{figure}[htbp]
\centering\includegraphics[width=\linewidth]{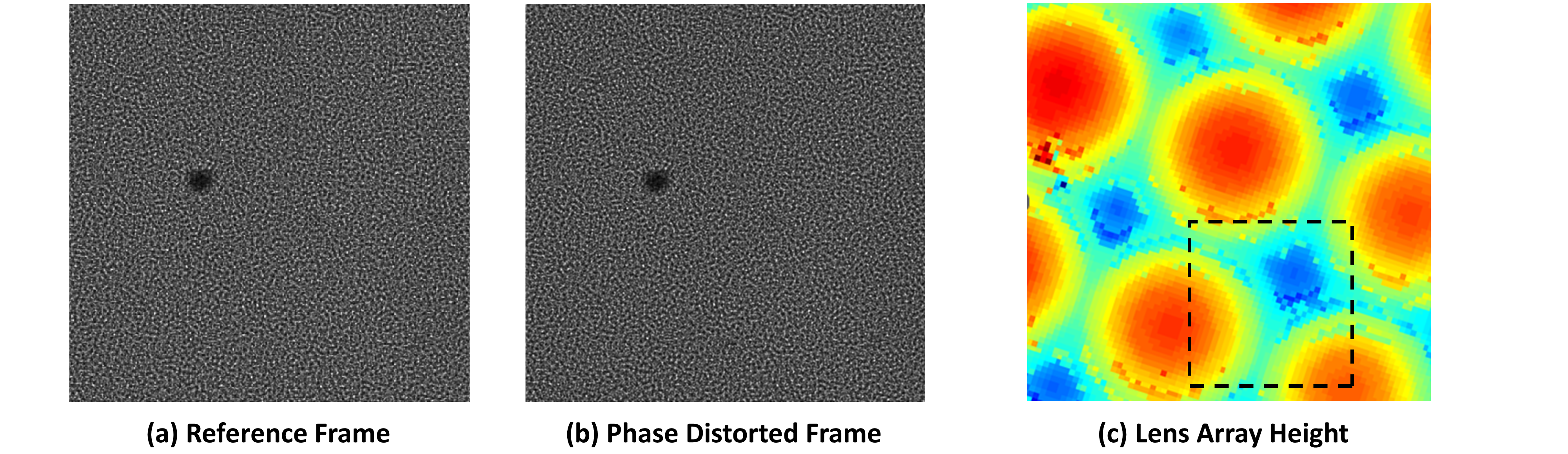}
\caption[Lens Array]{Wavefront sensor captured frames of lens array distortion. }
\label{PIV2Fig:WFS_lens}
\end{figure}

The coded wavefront sensor is also a variant of the classic phase
distortion problem~\cite{wang2018megapixel,wang2019quantitative}. The
principle is also similar to what is explained in
Section~\ref{PIV2Sec:PhasePrin}. The distortion is related to the
gradient of phase. A mask is placed in the front of the camera. A
frame without any distortion is captured as a reference frame
(Fig~\ref{PIV2Fig:WFS_lens} (a)). After the phase lens array
(Fig~\ref{PIV2Fig:WFS_lens} (c)) causing distortion, phase distorted
image (Fig~\ref{PIV2Fig:WFS_lens} (b)) is
captured. Fig~\ref{PIV2Fig:WFS_lens} (c) is a Zygo (interferometric)
measurement of the lens array, where the data was provided by the
authors of~\cite{wang2019quantitative}.

\begin{figure}[htbp]
\centering\includegraphics[width=\linewidth]{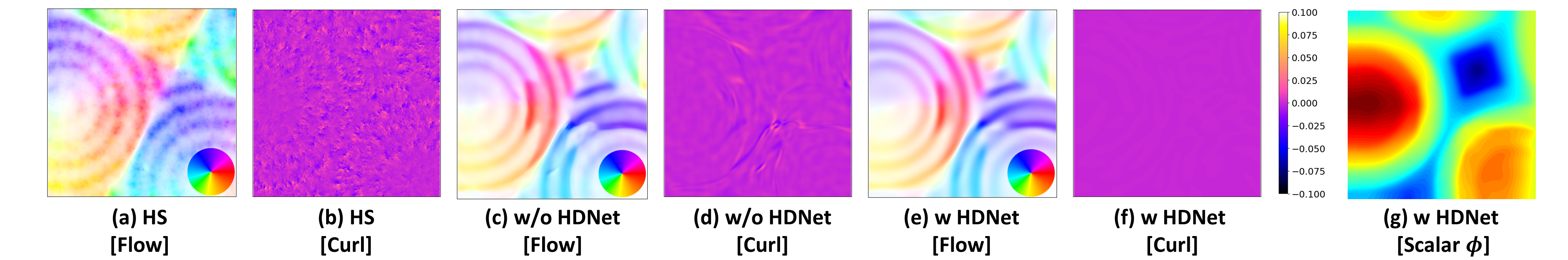}
\caption[WFS]{Wavefront sensor lens array reconstruction. This is the reconstruction of lens array. The lens array height is like Fig.~\ref{PIV2Fig:WFS_lens} (b). It is a regular array of circular lenses  separated by symmetrical gaps. (f) is the lens array phase reconstruct.}
\label{PIV2Fig:WFS}
\end{figure}

\begin{figure}[htbp]
\centering\includegraphics[width=\linewidth]{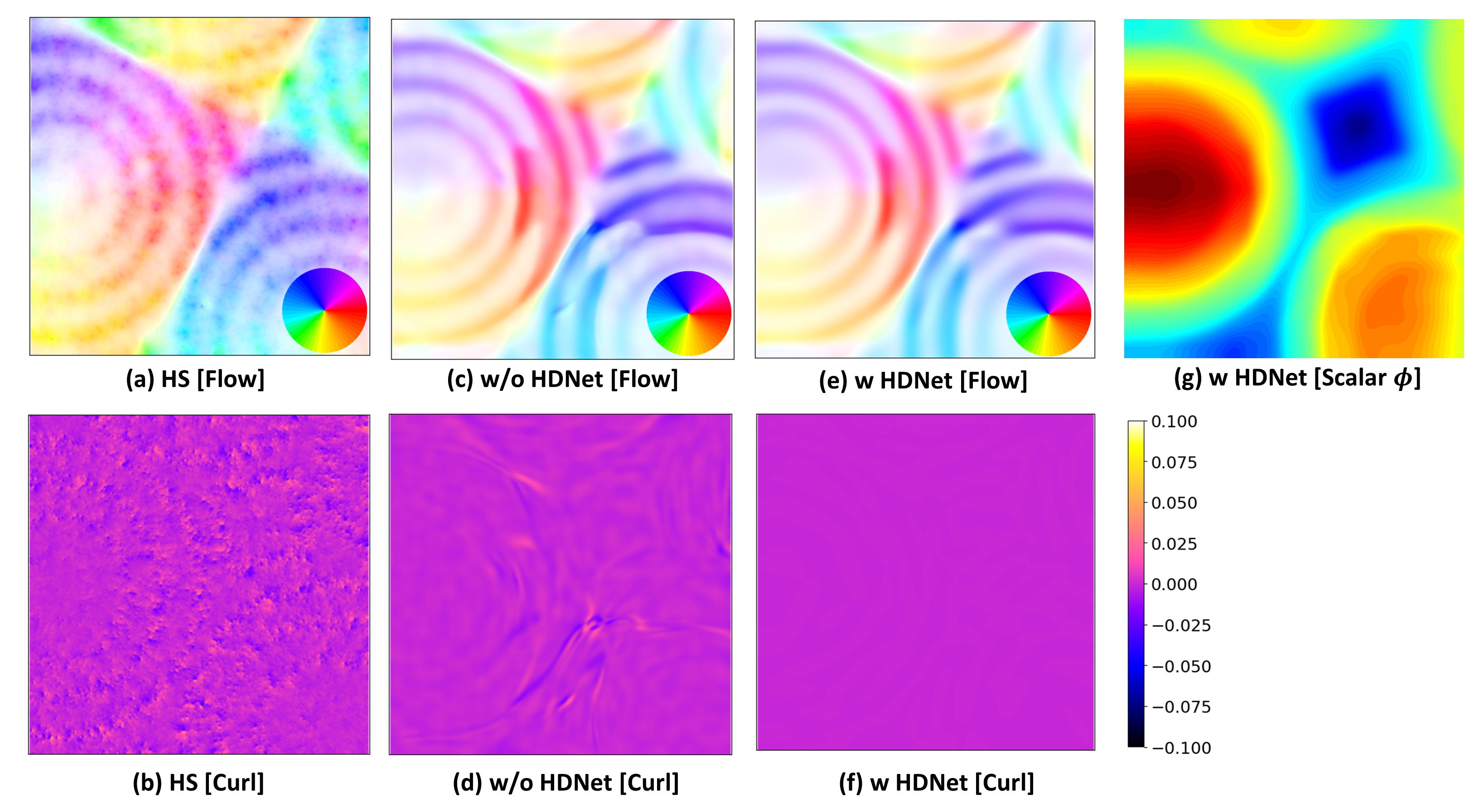}
\caption[WFS large]{Large figure version of Fig.~\ref{PIV2Fig:WFS} wavefront sensor lens array reconstruction. This is the reconstruction of lens array. The lens array height is like Fig.~\ref{PIV2Fig:WFS_lens} (b). It is a regular array of circular lenses  separated by symmetrical gaps. (f) is the lens array phase reconstruct.}
\label{PIV2Fig:WFS_large}
\end{figure}

The reconstruction in Fig.~\ref{PIV2Fig:WFS} is a crop of the lens
array as shown in the black dash square in the
Fig.~\ref{PIV2Fig:WFS_lens} (c). In the unconstrained optical flow
measurement we can see that the flow estimates contain erroneous
stair-stepping which is not present in the high accuracy
interferometric measurements.  Comparing (a),(c),(e), we can see that
our method has better flow reconstruction quality and fewer
artifacts. Comparing (b),(d),(f), we can see that our method have
physical constraint performance and curl error value is close to
zero. (g) is the phase reconstruction of our method. It is HDNet
scalar output. We can see it is symmetrical which match with but have
better reconstruction quality than the Zygo measurement in
Fig~\ref{PIV2Fig:WFS_lens} (c).

\subsection{PIV Synethetic Data}\label{PIV2Sec:PIVSynPar}

For the PIV simulations, we first generated 1 frame with 10000
particles in random position with pixels size about 1-2 pixels. The
particle pixels value was generated in a normal distribution with mean
1 and variance 0.2. The particle frame figure is shown in Supplement
Fig.~\ref{PIV2Fig:SyntheticParticle}. Then, we used the flow from
Helmholtz Synthesis inference dataset to warp the particle to get the
other particle image sequence.

\begin{figure}[htbp]
\centering\includegraphics[width=\linewidth]{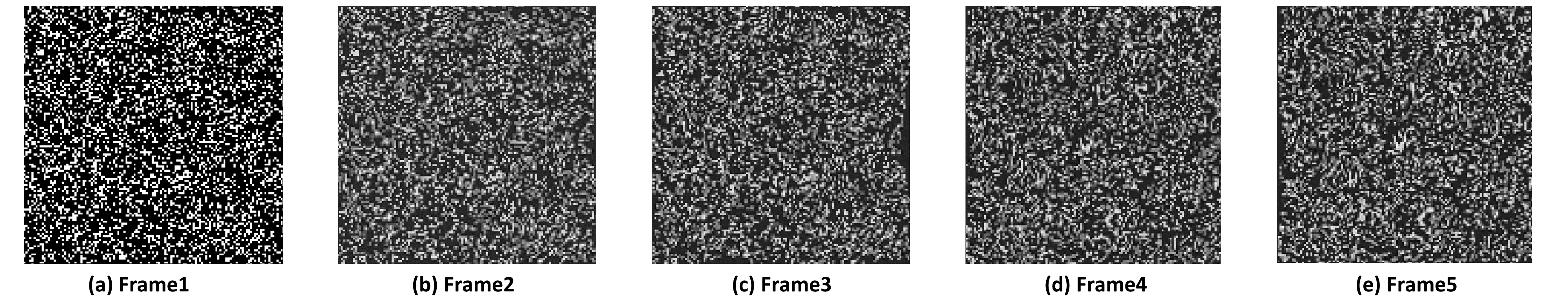}
\caption[Synthetic PIV]{Synthetic PIV particle image sequence $\ReflectiveIndex_t$}
\label{PIV2Fig:SyntheticParticle}
\end{figure}

\begin{figure}[htbp]
\centering\includegraphics[width=\linewidth]{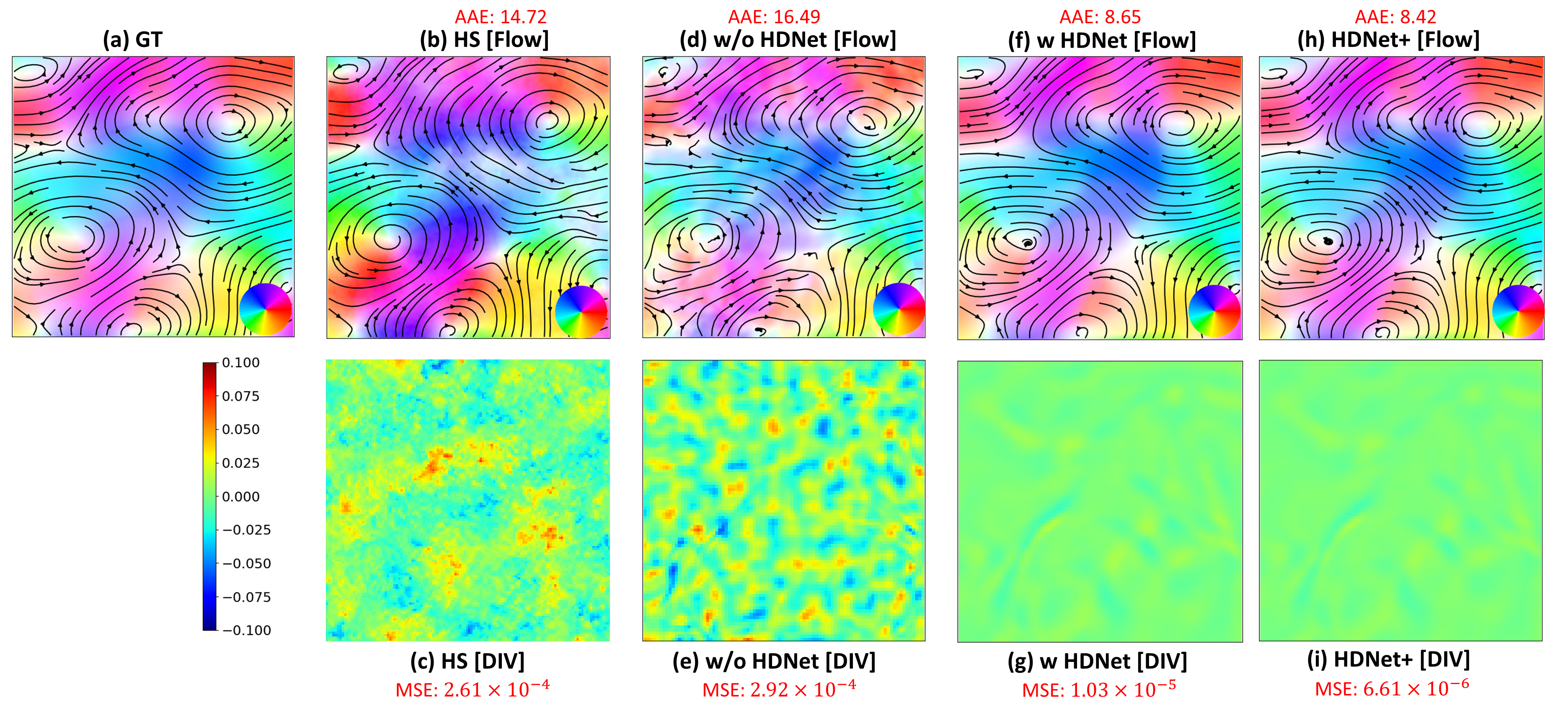}
\caption[Synthetic PIV large]{Large figure version of Fig.~\ref{PIV2Fig:SyntheticPIV} synthetic PIV data reconstruction comparison. (h) and (i) are the flow and divergence for HDNet+ which is the method that add a divergence penalty term to the total loss to the HDNet flow reconstruction. }
\label{PIV2Fig:SyntheticParticle_large}
\end{figure}

\paragraph{HDNet+}  It is very straightforward to add a divergence penalty term to the total loss for the flow reconstruction pipeline. To compare with different settings and show the flexibility of HDNet, we also did a comparison experiment that adds a divergence penalty to the total loss for our HDNet flow reconstruction pipeline to explore better reconstruction quality and physical constraint performance. The results are shown in Fig.~\ref{PIV2Fig:SyntheticParticle_large} (h),(i). The reconstruction quality and physical constraint performance are a little bit better.

\subsection{PIV Real Experiment Data}\label{PIV2Sec:PIVRealPar}

The real PIV experiment data is provided to us by [omitted for
  anonymity]. It is captured with a Phantom2640 camera with a
resolution of $1024 \times 976$, an exposure time of $99.540~\mu s$,
and a frame rate of $10000~fps$. We cropped the image to have a $256
\times 256$ size. Flow particle size is $10~\mu m$.

\begin{figure}[htbp]
\centering\includegraphics[width=\linewidth]{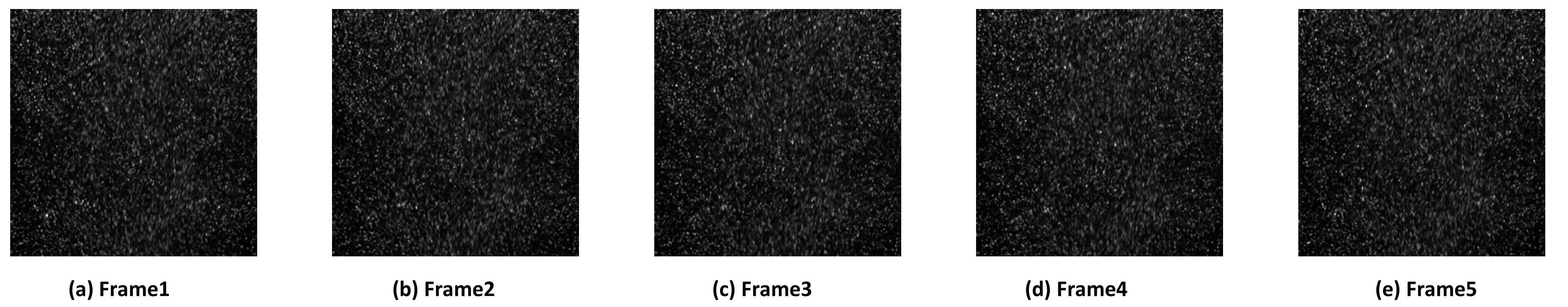}
\caption[Real Experiment PIV]{Real experiment PIV particle image sequence $\ReflectiveIndex_t$}
\label{PIV2Fig:RealParticle}
\end{figure}

\begin{figure}[htbp]
\centering\includegraphics[width=\linewidth]{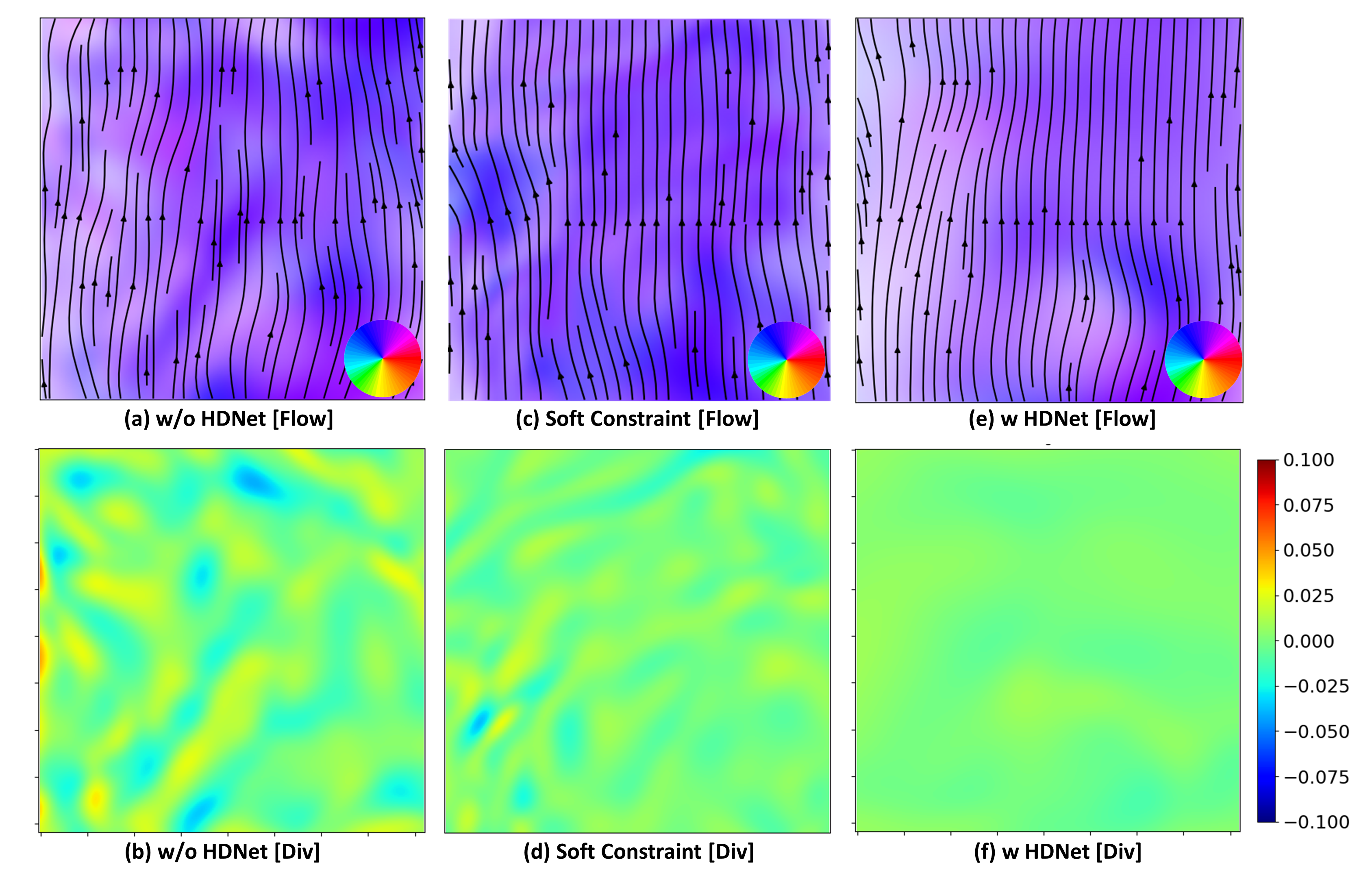}
\caption[Real PIV experiment Large]{Large figure version of Fig.~\ref{PIV2Fig:RealPIV} real PIV data reconstruction comparison. Soft constraint is the method that add a divergence penalty term to the total loss.}
\label{PIV2Fig:RealParticle_large}
\end{figure}

\subsection{BOS Experiment Data}
The data for the BOS experiment is taken
from~\cite{atcheson2009evaluation} and was provided to us by the
authors.  The target air distortion for reconstruction is the hot air
plume of a burning candle. Image resolution in this case is $270
\times 270$. The reconstructed region corresponds to the upper area of
the candle hot air. The distortion in these datasets is very small and
often only introduces subpixel shifts in the images. The dataset is
also particularly challenging since it uses cameras that record a
compressed video stream, so that MPEG artifacts further alter the
small distortions. The results in the main paper show that HDNet can
provide crucial physical regularization to this very difficult inverse
problem.

\begin{figure}[htbp]
\centering\includegraphics[width=\linewidth]{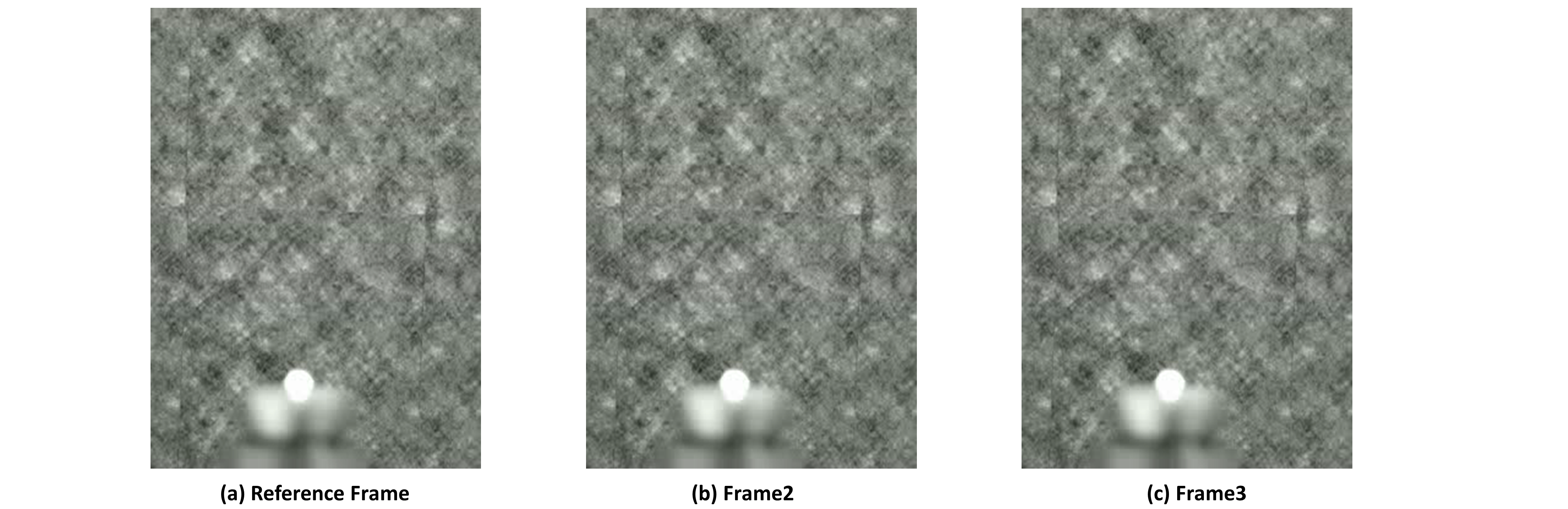}
\caption[BOS]{Real experiment BOS image sequence $\ReflectiveIndex_t$. This is gas turbulence on a burning candle. The distortion is very tiny. }
\label{PIV2Fig:BOSimgs}
\end{figure}

\begin{figure}[htbp]
\centering\includegraphics[width=\linewidth]{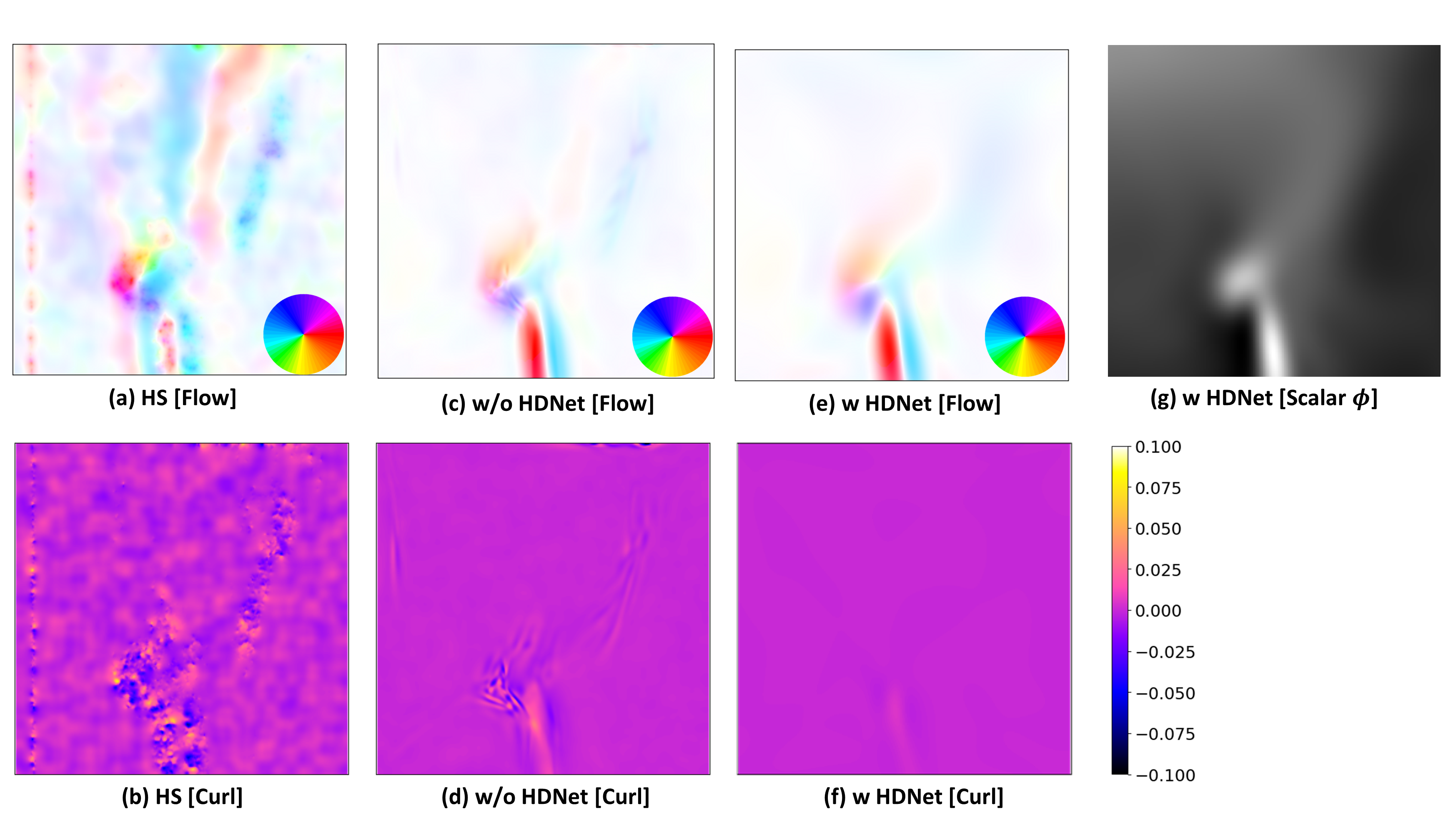}
\caption[BOS large]{Large figure version of Fig.~\ref{PIV2Fig:SchilierenImaging} BOS reconstruction comparison. Curl mean  the curl value map of left figure. w HDNet scalar $\ScalarPotential$ means our flow reconstruction pipeline with HDNet and the scalar output of HDNet, that clearly shows the hot air plume above the candle.}
\label{PIV2Fig:BOSimgs_large}
\end{figure}

\section{Wavelet Implicit Neural Representation} \label{PIV2Sec:WIRE}
Study reveals that directly learning the image or 2D/3D field with MLP leads to very poor accuracy~\cite{tancik2020fourier}.  One reason is that only the MLP can not learn high frequency of the image. Employing Gabor wavelet as the activation function can enable the MLP to learn the high frequency of the image~\cite{saragadam2023wire}. Every layer in MLP can be expressed as:
\begin{equation}
\mathbf{y_m}=\sigma(W_m \mathbf{y_{m-1}+b_m}),
\end{equation}  

where $W_m, \mathbf{b_m}$ are the weight and bias for the m layers~\cite{saragadam2023wire}; $\sigma$ is the activation function.

\begin{equation}
\sigma(x)=e^{j\frepara x}e^{-|s_0x|^2}
\end{equation}  

$\frepara$ determine the frequency level that it represents (Supplement Fig.~\ref{PIV2Fig:WIRE}). A smaller $\frepara$ generates smoother results corresponding to the ``coarse" reconstruction. A large $\frepara$ generates more high-frequency detail corresponding to ``fine" reconstruction. By using this property, we can achieve the coarse-to-fine reconstruction which will be discussed in Supplement Section~\ref{PIV2Sec:coarse-to-fine}. 

\textbf{Adaptive Learnable Parameter} Moreover, the WIRE representation exhibits adaptability, as its representation parameters, $\frepara$ and $\sigma_0$, are learnable according to the characteristics of the scene being represented. Comparing with NeRF position encoding neural representation, WIRE neural presentation is more continuous for changing the parameter. WIRE is faster than the Fourier Feature~\cite{tancik2020fourier} and robust for inverse problems of images and video~\cite{saragadam2023wire}.

\section{Frequency Coarse to Fine}\label{PIV2Sec:coarse-to-fine}

\begin{figure}[htbp]
\centering\includegraphics[width=0.8\linewidth]{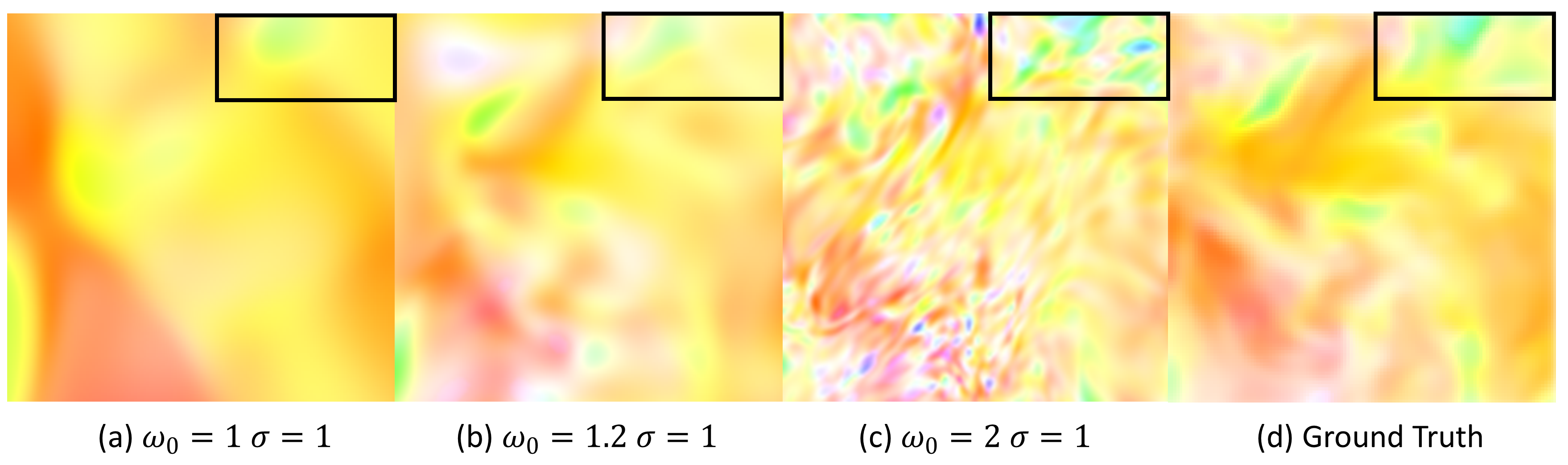}
\caption[WIRE Representation]{This is the WIRE representation flow reconstruction result with different $\frepara$ and $s_0$. A smaller $\frepara$ generates smoother results corresponding to the ``coarse" reconstruction. A large $\frepara$ generates more high-frequency detail corresponding to ``fine" reconstruction. }
\label{PIV2Fig:WIRE}
\end{figure}

\begin{figure}[htbp]
\centering\includegraphics[width=0.5\linewidth]{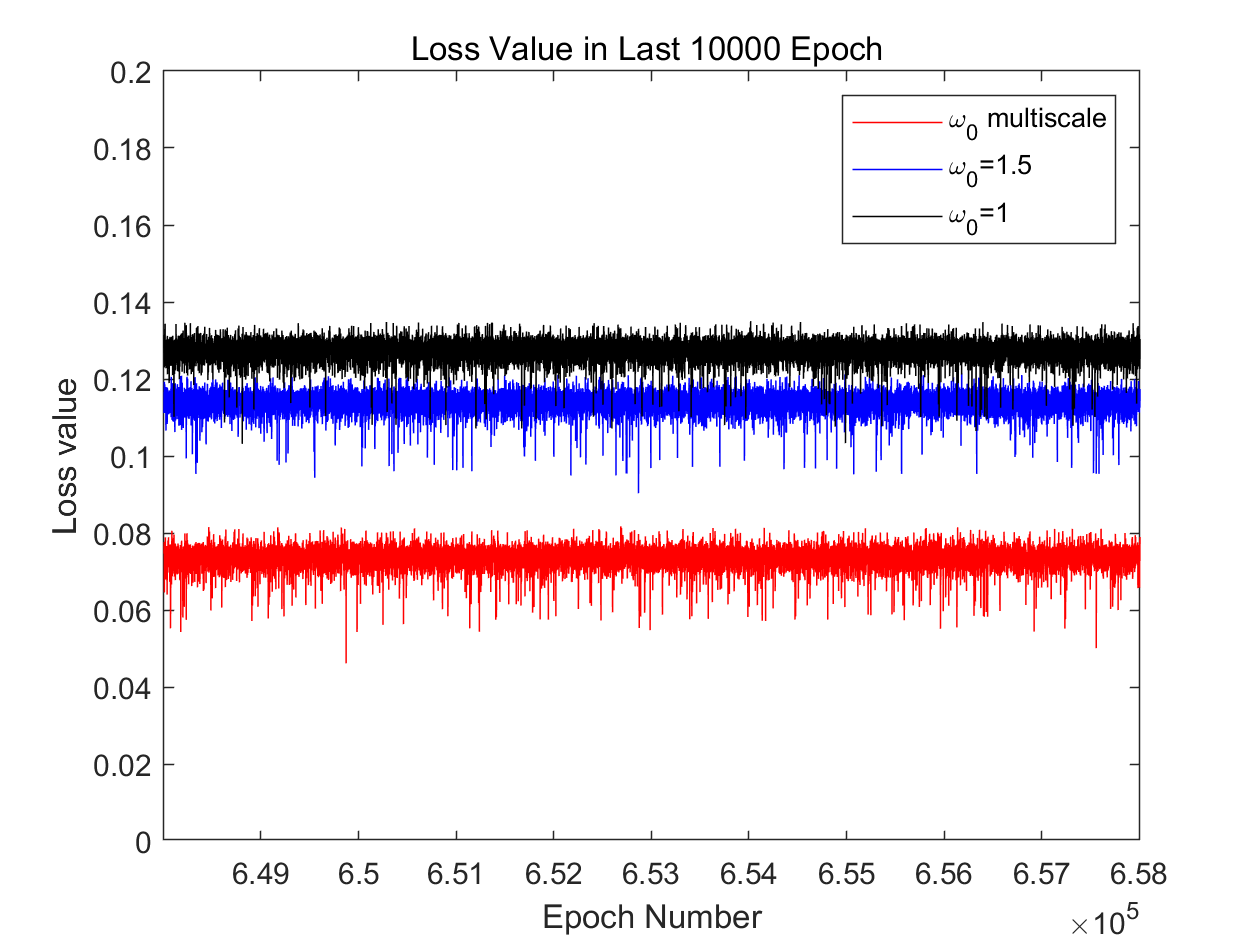}
\caption[Loss Figure]{Different setup loss figure. We can observe that with a single $w_0$ frequency the loss value in the end is always stagnant in a high value. This means the solution is trapped in a local minima. After employing the coarse-to-fine multiscale strategy, the loss decreases to a lower value, which means our coarse-to-fine can overcome the local minima.}
\label{PIV2Fig:lossfig}
\end{figure}

There is always a trade-off between achieving accuracy at the local and global levels in flow motion reconstruction. For example, as shown in Supplement Fig.~\ref{PIV2Fig:WIRE}, if $w_0$ is small, such as $w_0=1.2$,  the reconstructed flow exhibits an accurate overall shape, but lack the detailed information as shown in Supplement Fig.~\ref{PIV2Fig:WIRE} (b) the black rectangular . Conversely, when $w_0$ is too large, such as $w_0=2$, there is detail, but the global flow is not correct. This phenomenon arises due to the presence of local minima in the training process for a specific frequency representation (see Supplement Fig.~\ref{PIV2Fig:lossfig} for more detail). To overcome this trade-off we propose a coarse-to-fine approach that starts with low frequency to high frequency.

As discussed in Supplement Section~\ref{PIV2Sec:WIRE}, small $w_0$ means coarse representation and larger $w_0$ means fine representation. This property is used to implement the coarse-to-fine strategy. We first start with small $w_0$ and then progressively increase $w_0$ to a large value as the epoch number increases.  

The last step, activating learnability for the parameter $w_0$. As explained in the Section~\ref{PIV2Sec:WIRE}, learnable parameter undergoes automatic fine adjustments based on the scene.

%%%%%%%%%%%%%%%%%%%%%%%%%%%%%%%%%%%%%%%%%%%%%%%%%%%%%%%%%%%%

\newpage
\section*{NeurIPS Paper Checklist}

\begin{enumerate}

\item {\bf Claims}
    \item[] Question: Do the main claims made in the abstract and introduction accurately reflect the paper's contributions and scope?
    \item[] Answer: \answerYes{} % Replace by \answerYes{}, \answerNo{}, or \answerNA{}.
    \item[] Justification: the abstract clearly states the
      contribution and describes the experimental results.

\item {\bf Limitations}
    \item[] Question: Does the paper discuss the limitations of the work performed by the authors?
    \item[] Answer: \answerYes{} % Replace by \answerYes{}, \answerNo{}, or \answerNA{}.
    \item[] Justification: a section on Limitations and Future Work
      has been included.

\item {\bf Theory Assumptions and Proofs}
    \item[] Question: For each theoretical result, does the paper provide the full set of assumptions and a complete (and correct) proof?
    \item[] Answer: \answerNA{} % Replace by \answerYes{}, \answerNo{}, or \answerNA{}.
    \item[] Justification:  the paper does not include theoretical results. 

    \item {\bf Experimental Result Reproducibility}
    \item[] Question: Does the paper fully disclose all the information needed to reproduce the main experimental results of the paper to the extent that it affects the main claims and/or conclusions of the paper (regardless of whether the code and data are provided or not)?
    \item[] Answer: \answerYes{} % Replace by \answerYes{}, \answerNo{}, or \answerNA{}.
    \item[] Justification: The method is fully described in the paper
      and supplement, including the data sources and the synthetic data
      generation. Full code will be provided with the final paper.

\item {\bf Open access to data and code}
    \item[] Question: Does the paper provide open access to the data and code, with sufficient instructions to faithfully reproduce the main experimental results, as described in supplemental material?
    \item[] Answer: \answerNo{} % Replace by \answerYes{}, \answerNo{}, or \answerNA{}.
    \item[] Justification: it was not possible to anonymize the code
      and data for public release before the deadline. Both will be
      released upon acceptance.  

\item {\bf Experimental Setting/Details}
    \item[] Question: Does the paper specify all the training and test details (e.g., data splits, hyperparameters, how they were chosen, type of optimizer, etc.) necessary to understand the results?
    \item[] Answer: \answerYes{} % Replace by \answerYes{}, \answerNo{}, or \answerNA{}.
    \item[] Justification: Training is described in the main paper
      with some additional details in the supplement. In addition all
      code will be provided after acceptance.

\item {\bf Experiment Statistical Significance}
    \item[] Question: Does the paper report error bars suitably and correctly defined or other appropriate information about the statistical significance of the experiments?
    \item[] Answer: \answerYes{} % Replace by \answerYes{}, \answerNo{}, or \answerNA{}.
    \item[] Justification: Mean and variance are provided for the
      HDNet itself. In addition we provide example of applications
      as individual case studies. These do not have large enough
      sample size to compute statistics.

\item {\bf Experiments Compute Resources}
    \item[] Question: For each experiment, does the paper provide sufficient information on the computer resources (type of compute workers, memory, time of execution) needed to reproduce the experiments?
    \item[] Answer: \answerYes{} % Replace by \answerYes{}, \answerNo{}, or \answerNA{}.
    \item[] Justification: The compute resources (single user
      workstation) are detailed in the paper.
    
\item {\bf Code Of Ethics}
    \item[] Question: Does the research conducted in the paper conform, in every respect, with the NeurIPS Code of Ethics \url{https://neurips.cc/public/EthicsGuidelines}?
    \item[] Answer: \answerYes{} % Replace by \answerYes{}, \answerNo{}, or \answerNA{}.
    \item[] Justification: all guidelines were followed.

\item {\bf Broader Impacts}
    \item[] Question: Does the paper discuss both potential positive societal impacts and negative societal impacts of the work performed?
    \item[] Answer: \answerNA{} % Replace by \answerYes{}, \answerNo{}, or \answerNA{}.
    \item[] Justification: flow estimation is a technical problem for
      many scientific and engineering tasks, but without broad
      societal impact.
    
\item {\bf Safeguards}
    \item[] Question: Does the paper describe safeguards that have been put in place for responsible release of data or models that have a high risk for misuse (e.g., pretrained language models, image generators, or scraped datasets)?
    \item[] Answer: \answerNA{} % Replace by \answerYes{}, \answerNo{}, or \answerNA{}.
    \item[] Justification: the work poses no risk for misuse.

\item {\bf Licenses for existing assets}
    \item[] Question: Are the creators or original owners of assets (e.g., code, data, models), used in the paper, properly credited and are the license and terms of use explicitly mentioned and properly respected?
    \item[] Answer: \answerYes{} % Replace by \answerYes{}, \answerNo{}, or \answerNA{}.
    \item[] Justification: data sources are provided and author
      permission has been obtained.

\item {\bf New Assets}
    \item[] Question: Are new assets introduced in the paper well documented and is the documentation provided alongside the assets?
    \item[] Answer: \answerYes{} % Replace by \answerYes{}, \answerNo{}, or \answerNA{}.
    \item[] Justification: the text contains a full description of the
      generation of the synthetic training data. Code will also be
      provided after acceptance.

\item {\bf Crowdsourcing and Research with Human Subjects}
    \item[] Question: For crowdsourcing experiments and research with human subjects, does the paper include the full text of instructions given to participants and screenshots, if applicable, as well as details about compensation (if any)? 
    \item[] Answer: \answerNA{} % Replace by \answerYes{}, \answerNo{}, or \answerNA{}.
    \item[] Justification:  the paper does not involve crowdsourcing nor research with human subjects.

\item {\bf Institutional Review Board (IRB) Approvals or Equivalent for Research with Human Subjects}
    \item[] Question: Does the paper describe potential risks incurred by study participants, whether such risks were disclosed to the subjects, and whether Institutional Review Board (IRB) approvals (or an equivalent approval/review based on the requirements of your country or institution) were obtained?
    \item[] Answer: \answerNA{} % Replace by \answerYes{}, \answerNo{}, or \answerNA{}.
    \item[] Justification: the paper does not involve crowdsourcing nor research with human subjects.
\end{enumerate}

\end{document}